\def\paperID{9503}
\def\confName{ICCV}
\def\confYear{2025}

\def\paperTitle{LightsOut: Diffusion-based Outpainting for Enhanced Lens Flare Removal}

\def\authorBlock{
    Shr-Ruei Tsai \quad
    Wei-Cheng Chang \quad
    Jie-Ying Lee \quad
    Chih-Hai Su \quad
    Yu-Lun Liu \vspace{0.5em} 
    \\
    \centerline{National Yang Ming Chiao Tung University} \vspace{0.5em}
}

\newif\ifreview 
\newif\ifarxiv \newcommand{\arxiv}{\arxivtrue}
\newif\ifcamera 
\newif\ifrebuttal 

\arxiv

\pdfoutput=1
\documentclass[10pt,twocolumn,letterpaper]{article}
\ifreview \usepackage[review]{cvpr} \fi
\ifarxiv \usepackage[pagenumbers]{cvpr} \fi
\ifrebuttal \usepackage[rebuttal]{cvpr} \fi
\ifcamera \usepackage{cvpr} \fi

\usepackage{graphicx}	
\usepackage{amsmath}	
\usepackage{amssymb}
\usepackage{algorithm}
\usepackage{algpseudocode}
\usepackage{bbding}
\usepackage{booktabs}
\usepackage{times}
\usepackage{microtype}
\usepackage{epsfig}
\usepackage{caption}
\usepackage{float}
\usepackage{pifont}
\usepackage{placeins}
\usepackage{color, colortbl}
\usepackage{stfloats}
\usepackage{enumitem}
\usepackage{tabularx}
\usepackage{xstring}
\usepackage{multirow}
\usepackage{xspace}
\usepackage{url}
\usepackage{subcaption}
\usepackage{xcolor}
\usepackage[hang,flushmargin]{footmisc}
\usepackage{soul}

\definecolor{tabfirst}{rgb}{0.96, 0.77, 0.77} 
\definecolor{tabsecond}{rgb}{0.98 , 0.93, 0.77} 

\ifcamera \usepackage[accsupp]{axessibility} \fi

% ...

\ifarxiv  \fi

\newcommand{\modify}[1]{{\textcolor{black}{#1}}}

\newcommand{\R}[1]{{%
    \textbf{%
        \ifstrequal{#1}{1}{\textcolor{red}{R#1}}{%
        \ifstrequal{#1}{2}{\textcolor{blue}{R#1}}{%
        \ifstrequal{#1}{3}{\textcolor{magenta}{R#1}}{%
        \ifstrequal{#1}{4}{\textcolor{teal}{R#1}}{%
                           \textcolor{cyan}{R#1}%
        }}}}%
    }%
}}

\usepackage{xr-hyper}

\makeatletter
\newcommand*{\addFileDependency}[1]{
  \typeout{(#1)}
  \@addtofilelist{#1}
  \IfFileExists{#1}{}{\typeout{No file #1.}}
}

\makeatother
\newcommand*{\myexternaldocument}[1]{
    \externaldocument{#1}
    \addFileDependency{#1.tex}
    \addFileDependency{#1.aux}
}

\definecolor{cvprblue}{rgb}{0.21,0.49,0.74}
\usepackage[pagebackref,breaklinks,colorlinks,allcolors=cvprblue]{hyperref}
\usepackage[capitalize]{cleveref}
\crefname{section}{Sec.}{Secs.}
\crefname{table}{Table}{Tables}
\crefname{figure}{Fig.}{Figs.}

\ifarxiv \crefname{appendix}{App.}{Apps.}
\else \crefname{appendix}{Suppl.}{Suppls.} \fi

\unless\ifarxiv \myexternaldocument{_supplementary} \fi

\begin{document}
\title{\paperTitle}
\author{\authorBlock}

\twocolumn[{%
\renewcommand\twocolumn[1][]{#1}%
\maketitle
\vspace{-7mm}
\begin{center}
\centering
\captionsetup{type=figure}
\resizebox{1.0\textwidth}{!} 
{
\includegraphics[width=\textwidth]{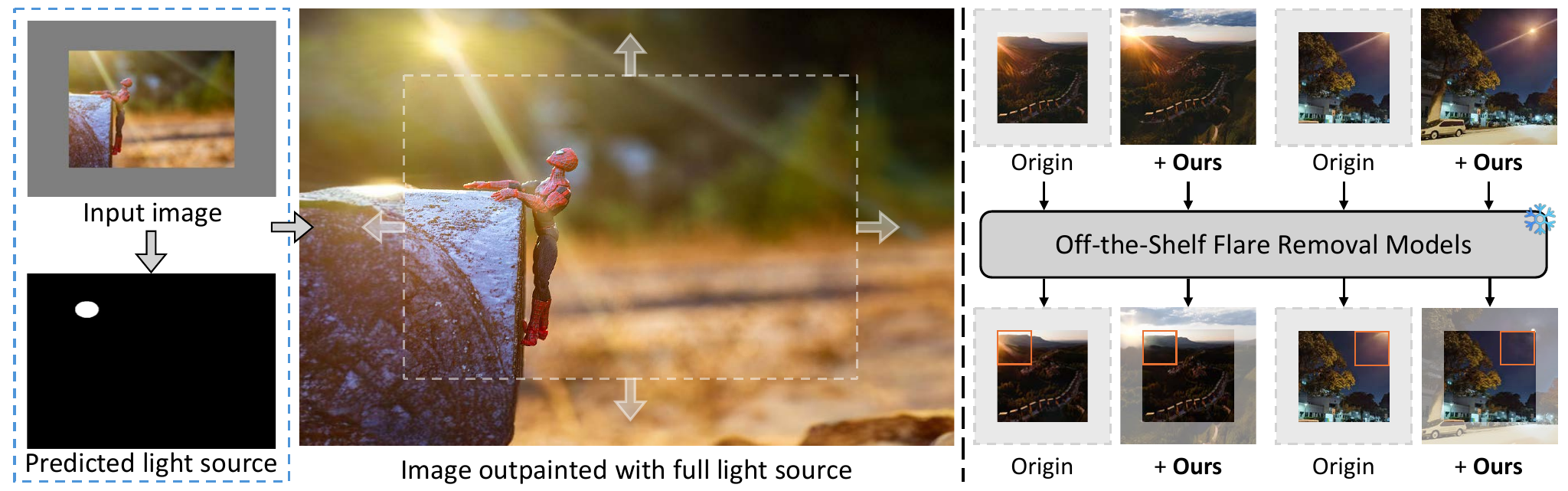}
}
\vspace{-7mm}
\caption{\textbf{Illustration of our diffusion-based outpainting method.}
Given an input image with incomplete or missing off-frame light sources, existing Single Image Flare Removal (SIFR) models struggle to effectively remove lens flare artifacts due to incomplete context. Our proposed approach accurately predicts and outpaints the off-frame light sources, allowing subsequent SIFR models to perform significantly better. As demonstrated, integrating our outpainting strategy as a plug-and-play preprocessing step substantially enhances flare removal quality and visual realism.}
\label{teaser}
\end{center}
}]

\maketitle

\begin{abstract}

Lens flare significantly degrades image quality, impacting critical computer vision tasks like object detection and autonomous driving. Recent Single Image Flare Removal (SIFR) methods perform poorly when off-frame light sources are incomplete or absent. We propose LightsOut, a diffusion-based outpainting framework tailored to enhance SIFR by reconstructing off-frame light sources. Our method leverages a multitask regression module and LoRA fine-tuned diffusion model to ensure realistic and physically consistent outpainting results. Comprehensive experiments demonstrate LightsOut consistently boosts the performance of existing SIFR methods across challenging scenarios without additional retraining, serving as a universally applicable plug-and-play preprocessing solution.
Project page: \url{https://ray-1026.github.io/lightsout/}

\end{abstract}
\section{Introduction}
\label{sec:intro}

Lens flare, categorized into reflective flares, scattering flares, and lens orbs (backscatter) \cite{li2021let, wu2021train, 10.1145/2010324.1965003}, significantly degrades image quality and negatively impacts computer vision tasks such as object detection and autonomous driving. Traditional flare removal methods \cite{asha2019auto, automated_removal, auto_removal2} relied on handcrafted cues like intensity thresholding or template matching but struggled with complex artifacts. Recent deep learning approaches, such as U-Net~\cite{wu2021train} and Uformer~\cite{dai2022flare7k}, have achieved substantial improvements due to dedicated datasets. Nevertheless, Single-Image Flare Removal (SIFR) remains challenging, particularly in nighttime scenarios, due to limited real-world paired training data.

Recent SIFR advances primarily focus on dataset construction and architectural improvements. Wu et al.\cite{wu2021train} introduced a flare dataset with captured and simulated images. Flare7k\cite{dai2022flare7k} and its enhanced version, Flare7k++\cite{dai2023flare7k++}, provide extensive synthetic and real flare data. Recent methods like Difflare\cite{zhou2024difflare} and MFDNet~\cite{jiang2024mfdnet} employ sophisticated architectures, such as diffusion models and multi-scale processing, achieving state-of-the-art performance.

\begin{figure}[t]
    \centering
    \includegraphics[width=\columnwidth]{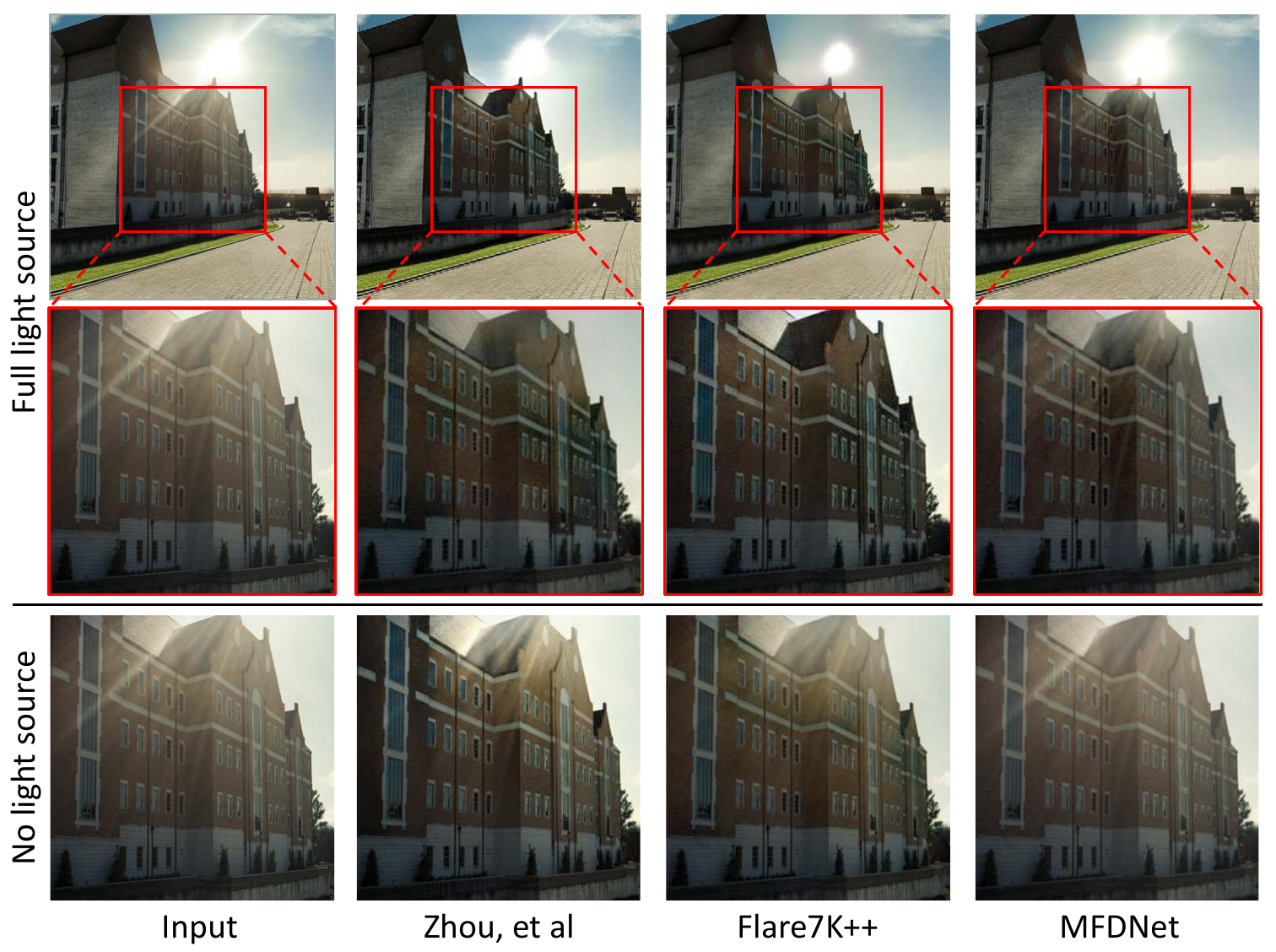}
    \vspace{-7mm}
    \caption{
    \textbf{Motivation for outpainting incomplete off-frame light sources.} (a) With complete off-frame light sources, state-of-the-art SIFR methods effectively remove lens flare artifacts. (b) In scenarios lacking complete views of off-frame sources, these methods degrade significantly, leaving noticeable artifacts. This highlights the importance of complete light source context, motivating our proposed outpainting solution.
    }
    \label{fig:why_figure}
    \vspace{-3mm}
\end{figure}

Despite these advancements, current SIFR methods still struggle when images lack complete views of off-frame light sources. As illustrated in \cref{fig:why_figure}, existing methods significantly degrade in such scenarios, resulting in increased flare artifacts and reduced realism. Metrics like PSNR and LPIPS~\cite{zhang2018unreasonable} confirm that complete light source information is crucial for effective flare removal.

To address this, we propose \textbf{LightsOut}, a diffusion-based outpainting method designed for accurate completion of missing off-frame light sources. Our approach integrates a multitask regression module to predict light source parameters precisely, alongside LoRA~\cite{hu2022lora} fine-tuning of a stable diffusion inpainting model explicitly conditioned on these predictions. This ensures that generated content aligns closely with real-world flare and illumination distributions.

As shown in~\cref{teaser}, our method seamlessly integrates as a plug-and-play preprocessing step with existing SIFR frameworks, significantly enhancing performance in challenging scenarios. Extensive quantitative and qualitative evaluations confirm that LightsOut consistently improves the realism and effectiveness of flare removal across various state-of-the-art methods.

Our contributions are summarized as follows:
\begin{itemize}
\item We identify and address a key limitation of SIFR methods dealing with incomplete off-frame light sources through a specialized decomposition strategy.
\item We propose a LoRA fine-tuned diffusion-based model that accurately reconstructs physically consistent off-frame light sources and flare artifacts.
\item We introduce a plug-and-play preprocessing framework that universally enhances existing SIFR models without additional retraining.
\end{itemize}

\vspace{-2pt}
\section{Related Work}
\label{sec:related}
\noindent {\bf Lens Flare Removal.}
Physical lens modifications \cite{asha2019auto} become ineffective with strong light sources \cite{dai2023nighttime}. Early approaches followed a flare detection-then-removal pipeline \cite{asha2019auto, automated_removal, auto_removal2}, focused on veiling glare removal \cite{10.5555/1208706, 10.1145/1276377.1276424} and reflective flare removal \cite{automated_removal, asha2019auto, auto_removal2, aerial_tracking, aerial_tracking}.
Deep learning approaches were limited by paired data scarcity. Wu et al. \cite{wu2021train} developed a dataset with a U-Net-based \cite{unet} approach. Dai et al. \cite{dai2022flare7k} expanded this with physics-based simulation \cite{10.1145/2010324.1965003, flare_simulation2}, later enhanced as Flare7K++ \cite{dai2023flare7k++}. Qiao et al. \cite{light_source} used unpaired data with a CycleGAN-inspired \cite{CycleGAN} framework. Recent architectures like MFDNet \cite{li2021let} and Zhou et al.'s approach \cite{zhou2022lednet} further improved performance.
Despite advances, current models degrade significantly when off-frame light sources are incomplete or absent, which is the limitation our work addresses.


\vspace{2pt}
\noindent {\bf Adapting Diffusion Models.}
Diffusion models \cite{ddpm, song2021scorebased, improvedddpm} have been adapted for image generation, editing, and restoration~\cite{rombach2022high, glide, palette, saharia2022photorealistic,yeh2024diffir2vr,hsiao2024ref,chen2024narcan,wang2024matting,chao2022denoising}. Early applications to inpainting by Sohl-Dickstein et al. \cite{sohl2015deep} and Song et al. \cite{song2021scorebased} showed promise. Guided synthesis approaches \cite{ilvr, sdedit} offered alternative conditioning strategies but have limitations for outpainting tasks.
Several approaches focus on fine-tuning pre-trained diffusion models~\cite{ruiz2023dreambooth, yang2023paint} or text embeddings~\cite{gal2022image, voynov2023p+} using a single or a few reference object images. 
Additionally, to improve fine-tuning efficiency, researchers have proposed methods that simplify the adaptation process of diffusion models~\cite{han2023svdiff, qiu2023controlling,hu2022lora}.
Low-Rank Adaptation (LoRA) \cite{hu2022lora} has emerged as an efficient fine-tuning approach. However, existing diffusion models lack physics-based modeling of optical phenomena \cite{10.1145/2010324.1965003, 1926JOSA...12..271H} and struggle with contextual extrapolation beyond image boundaries.

\vspace{2pt}
\noindent {\bf Image Completion and Outpainting.}
Traditional image completion used low-level cues \cite{Bertalmio2000image, ballester2001filling, bertalmio2003simultaneous, hays2007scene}. GAN-based methods \cite{gan, pathak2016context} introduced encoder-decoder architectures with innovations including dilated convolutions \cite{10.1145/3072959.3073659, yu2015multi}, partial/gated convolutions \cite{liu2018image, yu2019free}, contextual attention \cite{yu2018generative}, edge maps \cite{nazeri2019edgeconnect, xiong2019foreground, xu2020e2i, guo2021image}, and semantic segmentation \cite{hong2018learning, ntavelis2020sesame}.
Outpainting techniques include semantic regeneration networks \cite{srn}, edge-guided models \cite{edgeout}, spiral generation \cite{spiral}, and RCT blocks with LSTM \cite{nsipo, lstm}. Recent transformer-based methods \cite{queryqtr, uformer} still struggle with physically consistent light sources and flare extrapolation.
Diffusion-based approaches \cite{rombach2022high, palette, sdedit, sindiff, compositional,wu2025aurafusion360,liu2025corrfill} demonstrate generative capabilities but lack explicit modeling of light sources and optical effects.

\vspace{2pt}
\noindent {\bf Image Conditioned Diffusion Models.}
Diffusion models excel in generative tasks but struggle with fine-grained controllability. ControlNet~\cite{zhang2023adding} and IP-Adapter~\cite{ye2023ip} enhance reference-based generation by integrating input image features, while T2I-Adapter~\cite{mou2024t2i} conditions on external modalities like sketches and keypoints. InstructPix2Pix~\cite{brooks2023instructpix2pix} enables explicit attribute control via user instructions, and methods~\cite{sdedit, lugmayr2022repaint} refine generation using stochastic differential equations. Building on these advances, our study leverages the conditional techniques by introducing light source constraints as a novel conditioning factor.

\begin{figure*}[t]
    \centering
\vspace{-4mm}
    \includegraphics[width=\linewidth]{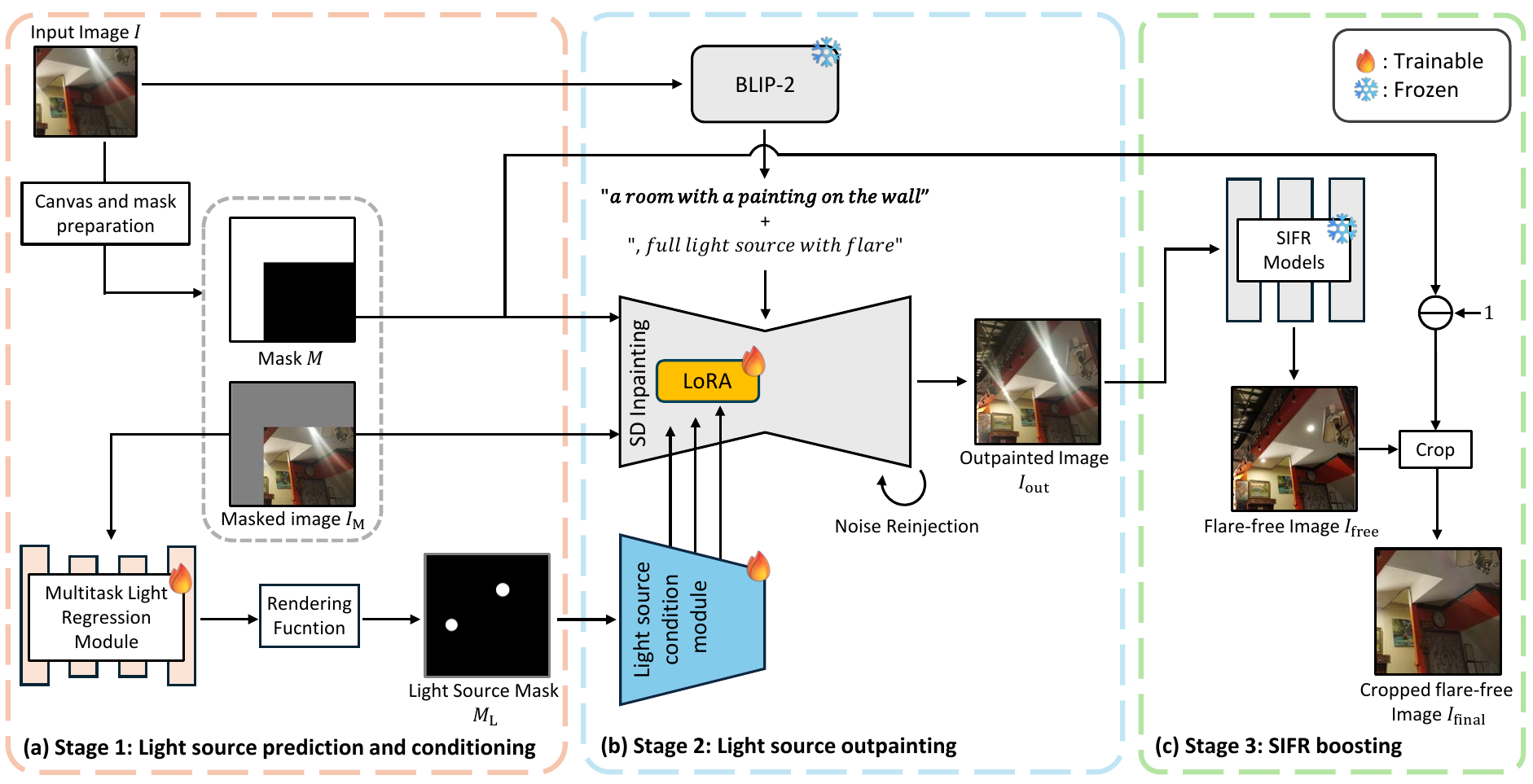}
\vspace{-7mm}
    \caption{\textbf{Overview of our proposed three-stage pipeline.}
    (a) \textbf{Light source prediction and conditioning}: We introduce a multitask regression module to accurately predict off-frame or incomplete light source parameters (positions, radii, and confidences). These predicted parameters guide a rendering function to generate the corresponding light source mask. (b) \textbf{Light source outpainting}: Leveraging a LoRA fine-tuned diffusion-based inpainting model with light source conditioning, our approach accurately outpaints both missing off-frame light sources and associated flare artifacts, producing visually coherent and realistic results. (c) \textbf{SIFR boosting}: Our generated outpainted images serve as enhanced inputs to existing SIFR methods, significantly improving their performance on previously challenging scenarios with incomplete light source information. The proposed pipeline thus effectively operates as a plug-and-play module to boost existing flare removal models.
    }
    \label{fig:pipeline}
\vspace{-4mm}
\end{figure*}

\section{Method}
\label{sec:method}
\vspace{-2pt}
Our objective is to resolve the issue of the degradation in existing SIFR models with limited light source information. As illustrated in \cref{fig:pipeline}, we propose \modify{a} three stage method: (a) Given a flare-corrupted input image $I_\text{in} \in \mathbb{R}^{H \times W \times 3}$ with an incomplete light source, we first define an outpainted region, producing the masked image $I_\text{M} \in \mathbb{R}^{H^\prime \times W^\prime \times 3}$, where $H^\prime > H$ and $W^\prime > W$,  and its corresponding binary mask $M \in \{0, 1\}^{H^\prime \times W^\prime \times 3}$. Then, we predict the light source mask $M_\text{L}$ for the guided condition (\cref{subsec:light_condition}). (b) After preparing the images and corresponding guided conditions, our outpainting approach completes the scene and reconstructs the full light source, constrained by $M_\text{L}$. The resulting outpainted image, denoted as $I_\text{out} \in \mathbb{R}^{H^\prime \times W^\prime \times 3}$, is generated as described in \cref{subsec:image_outpaint}. (c) The generated image $I_\text{out}$ is then processed by a SIFR model, yielding the flare-free image $I_\text{free} \in \mathbb{R}^{H^\prime \times W^\prime \times 3}$. Finally, we extract the origin region in $I_\text{in}$ from $I_\text{free}$ to get the final flare-free image $I_\text{final} \in \mathbb{R}^{H\times W\times 3}$ (\cref{subsec:flare_removal}). This three-stage pipeline effectively addresses the limitations of existing SIFR models by enhancing light source reconstruction before flare removal.

\vspace{-2pt}
\subsection{Preliminaries}
\vspace{-2pt}
\noindent{\bf Diffusion models.} transform a simple Gaussian noise distribution into the target data distribution. During the training phase, these models progressively add Gaussian noise to the original data $x_0$. Formally, at each diffusion step $t$, the noisy sample $x_t$ can be written as $x_t = \sqrt{\alpha_t}x_0 + \sqrt{1-\alpha_t}\epsilon,$ where $\epsilon \sim \mathcal{N}(0, \mathbf{I})$ is a Gaussian noise and $\alpha_t$ is a variance-scheduling parameter that governs how much noise is added. A neural network $\epsilon_{\theta}$ then learns to predict noise $\epsilon$ from the noisy sample $x_t$ by minimizing: 
$\mathcal{L} = \mathbb{E}_{x,t,\epsilon} \begin{Vmatrix}
        \epsilon_{\theta}(x_t, t, c) - \epsilon
    \end{Vmatrix}^2_2.$
where $c$ denotes conditioning signal such as text or masked images. At \modify{the} inference stage, the model starts from random noise and iteratively denoises the sample until it converges to a data point in the target distribution.

\vspace{2pt}
\noindent{\bf LoRA fine-tuning.} Instead of updating all parameters of the weight matrix \modify{$\mathcal{W} \in \mathbb{R}^{m \times n}$} \modify{in the denoising U-Net}, LoRA introduces a low-rank decomposition by injecting a small trainable matrix $\Delta W_i = A_i Bi$, where $A_i \in \mathbb{R}^{m\times d}, B_i \in \mathbb{R}^{d \times n}$, with $d \ll n$. The final weight matrix can be written as \modify{$\mathcal{W}^{\prime}_i = \mathcal{W}_i + \Delta \mathcal{W}_i$}, where only the added \modify{$\Delta \mathcal{W}$} is optimized during training, while the original weights \modify{$\mathcal{W}_i$} remain frozen.

\vspace{-2pt}
\subsection{Image Outpainting for Light Source}
\label{subsec:image_outpaint}
\vspace{-2pt}
Existing diffusion-based outpainting models often lack the task-specific adaptation necessary for high-quality light source reconstruction. To address this, we fine-tune the pre-trained Stable Diffusion v2 inpainting model, enabling realistic and structurally consistent light source generation.

\vspace{2pt}
\noindent{\bf Training.} To generate the outpainted results with complete light source, we inject LoRA weights and fine-tune them on the given $I_\text{M}$ and $M$. The loss function is
\begin{equation}
    \mathcal{L} = \mathbb{E}_{x, t, \epsilon, m} \begin{Vmatrix}
        \epsilon_{\theta}(x_t, t, p, M, I_\text{M}) - \epsilon
    \end{Vmatrix}^2_2,
\end{equation}
where $x \in I_\text{M}$, $p$ represents a text prompt derived from the input image using BLIP-2~\cite{li2023blip}, and $\odot$ denotes the element-wise product. 

\vspace{2pt}
\noindent{\bf Inference.} After training, the diffusion-based outpainting model aims to predict missing pixels at the corners of the masked region while preserving the integrity of the existing regions in $I_\text{M}$. One approach is ensuring that the masked regions are modified by incorporating the intermediate noisy state of the source data from the corresponding timestep in the forward diffusion process. This can be formulated as follows: 
\begin{equation}
    \begin{aligned}
        &x^\text{masked}_{t-1} = \sqrt{\bar{\alpha}_t} x_0 + \sqrt{1 - \bar{\alpha}_t} \epsilon, 
        \quad \epsilon \sim \mathcal{N}(0, \mathbf{I}), \\
        &x^\text{unmasked}_{t-1} = \mu_\theta(x_t, t) + \sigma_\theta(x_t, t) \cdot \epsilon, 
        \quad \epsilon \sim \mathcal{N}(0, \mathbf{I}), \\
        &x_{t-1} = M \odot x^\text{masked}_{t-1} + (1 - M) \odot x^\text{unmasked}_{t-1}.
    \end{aligned}
    \label{eq:laten_blend}
\end{equation}
where $\mu_\theta(x_t, t)$ and $\sigma_\theta(x_t, t)$ are the predicted mean and variance from the denoising model. Since our method builds on the Stable Diffusion inpainting model, the operations in \cref{eq:laten_blend} are performed in latent space. However, as observed in prior works \cite{zhu2023designing, tang2024realfill}, this can still introduce distortions in preserved regions of $I_\text{M}$, resulting in \modify{inconsistencies} in the generated output $I_\text{out}.$ To resolve this, instead of combining in latent space, we use the mask $M$ to perform alpha composition between $I_\text{out}$ and $I_\text{M}$ directly in the RGB space. It ensures that $I_\text{out}$ can be with full recovery on the existing area and a smooth transition at the boundary of the generated region.

\vspace{2pt}
\noindent{\bf Noise reinjection.} Although we composite in RGB space, noticeable discrepancies arise between masked and unmasked regions in the final output.  As described in \cref{eq:laten_blend}, the denoising steps treat the masked and unmasked regions as separate entities. This can cause inconsistencies and error accumulation across denoising steps. To mitigate this, we adopt noise reinjection, inspired by prior works~\cite{lugmayr2022repaint, wang2024your}, as formalized in \cref{algo:noise_reinjection}. This technique reintroduces noise at intermediate steps, allowing the model to re-denoise and better align with the correct distribution.
\begin{algorithm}[h]
\small
\caption{Noise reinjection}
\label{algo:noise_reinjection}

\begin{algorithmic}[1]
    \Require Masked image $I_\text{M}$, binary mask $M$, Pretrained SD inpainting model $\boldsymbol{\epsilon}_{\theta}$, Timesteps sequence $\{t_1, t_2, \dots, t_N\}$, and Repeat time $R$
    \Ensure Outpainted image $I_\text{out}$
    \State $r \gets R$
    \State $x_T \sim \mathcal{N}(0, \mathbf{I})$
    \State $i \gets N-1$
    \While{$i \geq 0$}
    \State $t \gets t_i$, $t_{prev} \gets t_{i+1}$
    \State $z_t \sim \mathcal{N}(0, \mathbf{I})$
    \State $x_{t-1} = \begin{aligned}[t]
                    &\sqrt{\bar{\alpha}_{t-1}} \left( \frac{x_t - \sqrt{1-\bar{\alpha}_t} \boldsymbol{\epsilon}_{\theta}(x_t, I_\text{M}, M, t)}{\sqrt{\bar{\alpha}_t}} \right) \\
                    &+ \sqrt{1 - \bar{\alpha}_{t_{prev}} - \sigma^2_t} \cdot \boldsymbol{\epsilon}_{\theta}(x_t, I_\text{M}, M, t) + \sigma_t z_t
                    \end{aligned}$
    \If {$i>0$ and $r>0$}
        \State $x_{t} \sim \mathcal{N}\left(\sqrt{\alpha_t}x_{t-1}, \sqrt{1-\alpha_t}\mathbf{I} \right)$    \Comment{Noise reinjection}
        \State $r \gets r-1$
    \Else
        \State $i \gets i-1$
        \State $r \gets R$
    \EndIf
    \EndWhile
    \State $I_\text{out} = x_0$
    \State \Return $I_\text{out}$
\end{algorithmic}
    
\end{algorithm}

\vspace{-2pt}
\subsection{Light Source Prediction and Conditioning}
\label{subsec:light_condition}
\vspace{-2pt}
Our outpainting approach (\cref{subsec:image_outpaint}) leverages Stable Diffusion’s generative capabilities to reconstruct scenes and their light sources. However, it faces limitations, including spatial misalignment, incomplete synthesis, and inconsistent handling of multiple light sources. To address these, we propose multitask regression and light source conditioning modules to enhance outpainting accuracy and realism.

\vspace{2pt}
\noindent{\bf Multitask regression module.}
We tackle the challenging task of predicting light sources within masked regions of $I_\text{M}$, which is more complex than prediction from complete images. Unlike conventional U-Net \cite{unet} methods generating full maps, we adopt a parameterized regression approach, modeling sources as circular entities to reduce computational cost, stabilize training, and ensure physically meaningful results.

\begin{figure}[t]
    \centering
    \includegraphics[width=\columnwidth]{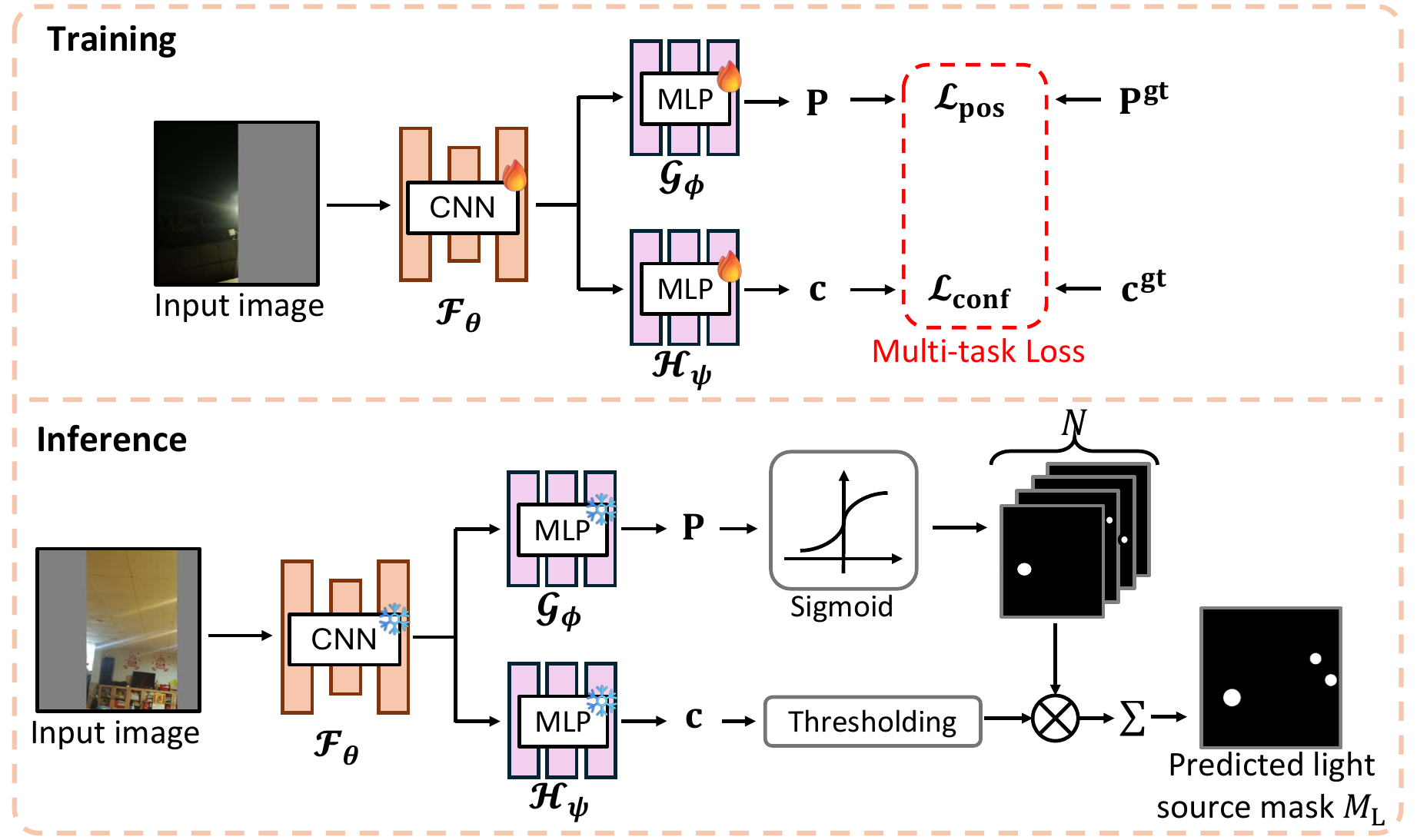}
\vspace{-7mm}
    \caption{\textbf{Overview of the multitask regression module.}
    Our model performs multitask regression to simultaneously predict two essential components: the physical parameters $\mathbf{P}$ and the corresponding confidence probabilities $\mathbf{c}$ for potential light sources. 
    During training, these are supervised by a designed multitask loss. At inference, predicted parameters are integrated to generate light source masks $M_\text{L}$.
    }
    \label{fig:lightsource_regression}
    \vspace{-3mm}
\end{figure}

The multitask regression module, as shown in \cref{fig:lightsource_regression}, predicts $N$ sets of $(x,y,r)$ parameters to environments with multiple light sources, where $(x,y)$ are the planar coordinates, $r$ is the radius, and $N$ serves as a hyperparameter. Since real-world scenes rarely contain exactly $N$ light sources, we introduce a confidence score to estimate the existence probability of each predicted source. The proposed architecture consists of a CNN-based feature extractor $\mathcal{F}_{\theta}$ and two specialized MLPs: $\mathcal{G}_{\phi}$ for estimating physical parameters and $\mathcal{H}_{\psi}$ for computing confidence scores. This architecture can be formally expressed as:
\begin{equation}
    \begin{bmatrix} \mathbf{P} \ \mathbf{c} \end{bmatrix} =
    \begin{bmatrix} \mathcal{G}{\phi}(\mathcal{F}{\theta}(I_{\text{tgt}})) \
    \mathcal{H}{\psi}(\mathcal{F}{\theta}(I_{\text{tgt}})) \end{bmatrix},
\end{equation}
where $\mathbf{P} = \begin{bmatrix} 
        x_1 & x_2 & \dots & x_\text{N} \\ 
        y_1 & y_2 & \dots & y_\text{N} \\ 
        r_1 & r_2 & \dots & r_\text{N} \\ 
    \end{bmatrix}^\top \in \mathbb{R}^{N \times 3}$ represents the matrix of predicted light source parameters, and $\mathbf{c} = \begin{bmatrix} 
        c_1 & c_2 & \dots & c_\text{N} 
    \end{bmatrix}^\top \in [0, 1]^{N\times 1}$ denotes the vector of confidence scores for each predicted light source. The optimization is designed with the position loss $\mathcal{L}_\text{pos}$:
\begin{equation}
    \mathcal{L}_\text{pos}(\mathbf{P}, \mathbf{P}_\text{gt}) = \sum_{i \in \{x,y,r\}} smooth_{L1}(\mathbf{P}-\mathbf{P}_\text{gt}),
\end{equation}
in which
\begin{equation}
    smooth_{L1}(x) = 
    \begin{cases}
        0.5x^2, & \text{if } |x|<1\\
        |x| -0.5 & \text{otherwise,}
    \end{cases}
\end{equation}
where $\mathbf{P}_\text{gt}$ represents the ground truth of the physical parameters. To ensure permutation invariance in the matching between predicted and ground-truth parameters, we adopt a bipartite matching strategy~\cite{carion2020end} to obtain the optimal assignment. Since our task involves predicting both the number and spatial distribution of light sources, we introduce an additional confidence loss $\mathcal{L}_\text{conf}$ to supervise the prediction of existence \modify{probabilities}. The confidence loss is formulated as the binary cross-entropy between the predicted $\mathbf{c}$ and the ground truth $\mathbf{c}_\text{gt}$:
\begin{equation}
    \mathcal{L}_\text{conf} = -\sum_i (\mathbf{c}_{\text{gt}, i}, \log \mathbf{c}_i + (1-\mathbf{c}_{\text{gt}, i})\log (1-\mathbf{c}_i)).
\end{equation}

To optimize only relevant predictions, we compute the position loss exclusively for pairs where $\mathbf{c}_\text{gt}=1$, ignoring cases where $\mathbf{c}_\text{gt}=0$ as these correspond to non-existent light sources. Furthermore, to address the inherent uncertainty in predicting both light source locations and their existence, we introduce an uncertainty-aware weighting mechanism\cite{kendall2018multi, liebel2018auxiliary}. Specifically, we model two learnable parameters, $\sigma_1$ and $\sigma_2$, and define the total loss function as:
\begin{equation}
    \mathcal{L} = \frac{1}{2\sigma^2_1}\mathcal{L}_\text{pos} + \frac{1}{2\sigma^2_2}\mathcal{L}_\text{conf} + \log(1+\sigma^2_1) + \log(1+\sigma^2_2)
\end{equation}
This allows the network to adaptively balance \modify{between} the losses, leading to more robust learning in either position or confidence predictions.

\vspace{2pt}
\noindent{\bf Rendering function.} During inference in the multitask regression module, we perform a forward pass to obtain the predicted $\mathbf{P}$ and $\mathbf{c}$. To generate a spatial representation of the light sources, we apply an activation function and a confidence threshold to suppress unreliable predictions, as illustrated in \cref{fig:lightsource_regression}. The final predicted \modify{light} source mask $M_\text{L}$ is computed as:
\begin{equation}
    M_\text{L}(x, y) = \sum^N_{i=1}\tilde{c}_i \cdot \sigma\left(r_i-\sqrt{(x-x_i)^2 + (y-y_i)^2} \right),
\end{equation}
where $\tilde{c}_i$ denotes the $\textbf{c}$ after thresholding, and $\sigma(\cdot)$ is the sigmoid function. The design of the function helps reconstruct the spatial representation by suppressing unreliable predictions.

\vspace{2pt}
\noindent{\bf Light source condition module} is designed to guide our outpainting approach mentioned in \cref{subsec:image_outpaint} by leveraging $M_\text{L}$, ensuring that light sources are generated in physically plausible locations. The module employs a learnable mechanism that conditions the generative process on the provided light source map, allowing explicit control over light placement. During optimization, we enforce an L2 loss:
\begin{equation}
    \mathcal{L}_\text{light} = \begin{Vmatrix}
        \tilde{M}_\text{L} - M_\text{L}
    \end{Vmatrix}^2_2,
\end{equation}
where $\tilde{M}_\text{L}$ represents the generated light source map. After training, we integrate the model into our outpainting model as constraints to effectively guide the generated content.

\vspace{-2pt}
\subsection{Flare Removal Methods}
\label{subsec:flare_removal}
\vspace{-2pt}
Our method reconstructs incomplete or off-frame light sources, providing accurate illumination context and enhancing flare removal effectiveness. It serves as a model-agnostic preprocessing step, seamlessly integrating into existing SIFR pipelines without architectural modifications.

\vspace{-2pt}
\section{Experiments}
\label{sec:exp}

\begin{table*}[t]
\centering
\footnotesize
\caption{
\textbf{Quantitative evaluation against state-of-the-art diffusion-based outpainting methods.}
We comprehensively compare our method with baseline and existing diffusion-based inpainting and outpainting approaches. Our solution demonstrates superior performance across diverse scenarios, validating the effectiveness of our diffusion-based strategy in enhancing subsequent flare removal tasks.
}
\label{tab:off_the_shelf}
\vspace{-3mm}
\resizebox{\textwidth}{!}{
\begin{tabular}{lll|cccccccccc}
\toprule
\multirow{2}{*}{{Setting}} & \multirow{2}{*}{{SIFR Model}} & \multirow{2}{*}{{Method}} & \multicolumn{5}{c}{{Flare7k Real}} & \multicolumn{5}{c}{{Flare7k Synthetic}} \\ \cmidrule(lr){4-8} \cmidrule(lr){9-13}
 &  &  & PSNR $\uparrow$ & SSIM $\uparrow$ & LPIPS $\downarrow$ & G-PSNR $\uparrow$ & S-PSNR $\uparrow$ & PSNR $\uparrow$ & SSIM $\uparrow$ & LPIPS $\downarrow$ & G-PSNR $\uparrow$ & S-PSNR $\uparrow$\\
\midrule
\multirow{15}{*}{\rotatebox{90}{No light source}}
& \multirow{5}{*}{Zhou et al.~\cite{zhou2023improving}}
   & Direct input & 26.42 & 0.8770 & \cellcolor{tabsecond}{0.0445} & 22.04 & 20.01 & \cellcolor{tabfirst}{31.86} & 0.9499 & \cellcolor{tabfirst}{0.0181}& \cellcolor{tabfirst}{25.16}& \cellcolor{tabsecond}{23.92} \\
 &   & SD-Inpainting~\cite{rombach2022high}   & \cellcolor{tabsecond}{26.84} & \cellcolor{tabsecond}{0.8823} & 0.0449 &
 \cellcolor{tabsecond}{22.73}& 20.04
 & 31.39  & \cellcolor{tabsecond}{0.9518} & \cellcolor{tabsecond}{0.0185} & 24.62 & \cellcolor{tabsecond}{23.92}\\
 &  & SDXL-Inpainting~\cite{podell2023sdxl} & 26.00 & 0.8774 & 0.0474 & 21.27 & 19.53  
 & 30.80 &  0.9506 &  0.0196 & 23.94 & 23.23 \\
 &  & PowerPaint~\cite{zhuang2024task}      
 & 26.59 & 0.8736 & 0.0559 & 22.66 & \cellcolor{tabsecond}{20.66} 
 & 28.90 & 0.9262 & 0.0456 & 23.32 & 22.72 \\
 &  & Ours            
 & \cellcolor{tabfirst}{27.09} & \cellcolor{tabfirst}{0.8856} & \cellcolor{tabfirst}{0.0424} & 
 \cellcolor{tabfirst}{23.07} & 
 \cellcolor{tabfirst}{21.12} &
 \cellcolor{tabsecond}{31.56}  & \cellcolor{tabfirst}{0.9534}  & \cellcolor{tabfirst}{0.0181}  &
 \cellcolor{tabsecond}{24.74}& 
 \cellcolor{tabfirst}{24.17}
 \\ \cmidrule{2-13}
& \multirow{5}{*}{Flare7k++~\cite{dai2023flare7k++}} 
   & Direct input     
   & 26.29 & 0.8337 & 0.0442
   & 21.35 & 18.71 
   & 31.28 & 0.9685 & 0.0151 
   &23.51 & 23.59\\
 &   & SD-Inpainting~\cite{rombach2022high}   & \cellcolor{tabsecond}{27.98}  & 
 \cellcolor{tabsecond}{0.8938} & 
 \cellcolor{tabsecond}{0.0421} & 
 \cellcolor{tabsecond}{23.63}  &
 \cellcolor{tabsecond}{21.12}  &
 \cellcolor{tabsecond}{33.43}  & 
 \cellcolor{tabsecond}{0.9704} & 
 \cellcolor{tabsecond}{0.0129} &
 \cellcolor{tabsecond}{25.86} &
   \cellcolor{tabsecond}{25.84}
 \\
 &  & SDXL-Inpainting~\cite{podell2023sdxl} 
 & 27.01 & 0.8893 & 0.0452 &22.13 &19.65 
 & 31.63 &  0.9675 &  0.0151 &23.87 &24.03 \\
 &  & PowerPaint~\cite{zhuang2024task}      
 & 27.10 & 0.8814 & 0.0839 & 22.20&20.92
 & 29.24 & 0.9289 &  0.0890&23.86 & 23.62\\
 &  & Ours            
 & \cellcolor{tabfirst}{28.41} &
 \cellcolor{tabfirst}{0.8956} & 
 \cellcolor{tabfirst}{0.0397} & 
 \cellcolor{tabfirst}{24.15} &
 \cellcolor{tabfirst}{22.83 }&
 \cellcolor{tabfirst}{33.91}  & 
 \cellcolor{tabfirst}{0.9719}  & 
 \cellcolor{tabfirst}{0.0120}   &
 \cellcolor{tabfirst}{26.24} &
 \cellcolor{tabfirst}{26.59}
 \\ \cmidrule{2-13}
& \multirow{5}{*}{MFDNet~\cite{jiang2024mfdnet}}
    & Direct input     
    & \cellcolor{tabsecond}{27.04} & \cellcolor{tabsecond}{0.8904} & \cellcolor{tabsecond}{0.0463} & 
    \cellcolor{tabsecond}{22.37}& 
    \cellcolor{tabsecond}{19.94} & 
    \cellcolor{tabsecond}{33.42} & \cellcolor{tabfirst}{0.9721} & \cellcolor{tabsecond}{0.0122}&
    \cellcolor{tabsecond}{26.43} &
    \cellcolor{tabfirst}{26.24}\\
 &   & SD-Inpainting~\cite{rombach2022high}   & 
 26.82 & 0.8886 & 0.0483 & 
 22.21&
 18.54 &
 32.42 &  0.9676 & 0.0145 &
 25.73 &24.93\\
 &  & SDXL-Inpainting~\cite{podell2023sdxl} &
 25.55 & 0.8816 & 0.0535 &
 21.23 & 16.66 &
 29.95 & 0.9605 & 0.0204 & 
 23.21 & 20.91 
 \\
 &  & PowerPaint~\cite{zhuang2024task}      
 & 25.28 & 0.8746 & 0.0509 & 20.49&17.11
 & 27.57 & 0.9181 & 0.0952 & 22.06& 20.61\\
 &  & Ours            & 
 \cellcolor{tabfirst}{27.43}  & 
 \cellcolor{tabfirst}{0.8940} & 
 \cellcolor{tabfirst}{0.0451} & 
 \cellcolor{tabfirst}{22.97}  & 
 \cellcolor{tabfirst}{20.49}  &
 \cellcolor{tabfirst}{33.54}  & 
 \cellcolor{tabsecond}{0.9714}& 
 \cellcolor{tabfirst}{0.0119} &
 \cellcolor{tabfirst}{26.53} & 
 \cellcolor{tabsecond}{25.89} \\
\midrule
\multirow{15}{*}{\rotatebox{90}{Incomplete light source}}
& \multirow{5}{*}{Zhou et al.~\cite{zhou2023improving}}
   & Direct input     
   & 26.05 & 0.8771 & 0.0480 &
   21.88 & 19.92&  
   30.03 & 0.9464 & 
   \cellcolor{tabsecond}{0.0210}&
   \cellcolor{tabfirst}{24.24} &23.14\\
 &    & SD-inpainting~\cite{rombach2022high}   & \cellcolor{tabfirst}{26.42} & \cellcolor{tabsecond}{0.8817} & \cellcolor{tabsecond}{0.0469} &
 \cellcolor{tabsecond}{22.59} &
 20.14 &
 \cellcolor{tabsecond}{30.07}  & \cellcolor{tabsecond}{0.9483}  & 0.0212&
 24.01 &
 \cellcolor{tabsecond}{23.43}\\
 &  & SDXL-Inpainting~\cite{podell2023sdxl} & 25.50 & 0.8773 & 0.0492 &
 21.34 & 19.59
 & 29.61 &  0.9466 &  0.0227
 & 23.35 & 23.12\\
 &  & PowerPaint~\cite{zhuang2024task}      & 25.87 & 0.8716 & 0.0595  &22.12
 &\cellcolor{tabsecond}{20.50}
 &  27.92 &  0.9227  &  0.0486 &22.73
 &22.34
 \\
 &  & Ours            & \cellcolor{tabsecond}{26.29} & \cellcolor{tabfirst}{0.8842} & \cellcolor{tabfirst}{0.0453} & 
 \cellcolor{tabfirst}{22.68} & 
 \cellcolor{tabfirst}{20.80} &
 \cellcolor{tabfirst}{30.11}  & \cellcolor{tabfirst}{0.9504} & \cellcolor{tabfirst}{0.0202} & \cellcolor{tabsecond}{24.06}
  &
 \cellcolor{tabfirst}{23.50}
 \\ \cmidrule{2-13}
& \multirow{5}{*}{Flare7k++~\cite{dai2023flare7k++}}
   & Direct input     
   & 26.07 & 0.8333 & 0.0463 &21.58 & 18.34&
   30.23 & 
   \cellcolor{tabsecond}{0.9672} &
   0.0160 &23.88 &23.70 \\
 &    & SD-Inpainting~\cite{rombach2022high}   & \cellcolor{tabsecond}{28.02} & \cellcolor{tabsecond}{0.8944} & \cellcolor{tabsecond}{0.0431} & 
 \cellcolor{tabsecond}{23.81} &
 \cellcolor{tabsecond}{21.71} &
 \cellcolor{tabsecond}{31.02}  &  
 0.9671  &
 \cellcolor{tabsecond}{0.0153}   &
 \cellcolor{tabsecond}{24.70} &
 \cellcolor{tabsecond}{25.06}
 \\
 &  & SDXL-Inpainting~\cite{podell2023sdxl} 
 & 26.99 & 0.8906 & 0.0453 &
  22.41 &20.12 
 & 30.33  & 0.9657 &  0.0164 &  23.73 &23.98\\
 
 &  & PowerPaint~\cite{zhuang2024task}      & 26.85 & 0.8802 & 0.0869 &22.94 &20.76
 & 28.00 &  0.9253 & 0.0923  &23.27 &23.12\\
 &  & Ours            
 & \cellcolor{tabfirst}{28.15} & \cellcolor{tabfirst}{0.8957} & \cellcolor{tabfirst}{0.0409} & 
 \cellcolor{tabfirst}{24.20} &
 \cellcolor{tabfirst}{22.24} & 
 \cellcolor{tabfirst}{31.38} &  \cellcolor{tabfirst}{0.9682} & \cellcolor{tabfirst}{0.0144} &
 \cellcolor{tabfirst}{25.01} & 
 \cellcolor{tabfirst}{25.57}
 \\ \cmidrule{2-13}
& \multirow{5}{*}{MFDNet~\cite{jiang2024mfdnet}}
    & Direct input     & \cellcolor{tabsecond}{26.53} & \cellcolor{tabsecond}{0.8886} & \cellcolor{tabfirst}{0.0457} & 
    \cellcolor{tabsecond}{22.07}& \cellcolor{tabsecond}{20.08} &
    \cellcolor{tabsecond}{31.52} & \cellcolor{tabfirst}{0.9701} & \cellcolor{tabsecond}{0.0137} &
    \cellcolor{tabsecond}{25.90} &
    \cellcolor{tabsecond}{25.45}
    \\
 &   & SD-Inpainting~\cite{rombach2022high}   & 26.48 & 0.8884 & 
 \cellcolor{tabsecond}{0.0496} &
 \cellcolor{tabsecond}{22.07} & 19.09  &
 31.32  & 
 0.9672 &
 0.0155 &
 25.49& 
 24.79\\
 &  & SDXL-Inpainting~\cite{podell2023sdxl}
 & 24.90 & 0.8807 & 0.0565 &
 20.78 &16.27&
 29.63 & 0.9605 &  0.0208 & 
  23.57 &22.27\\
 &  & PowerPaint~\cite{zhuang2024task}  & 24.88 & 0.8728 & 0.0535 &20.10 &16.66 & 26.89 & 0.9156 & 0.0982 & 21.87&20.48\\
 
 &  & Ours            &
 \cellcolor{tabfirst}{26.94} & \cellcolor{tabfirst}{0.8922} & \cellcolor{tabfirst}{0.0457} & 
 \cellcolor{tabfirst}{22.75} &
 \cellcolor{tabfirst}{20.43} &
 \cellcolor{tabfirst}{31.60} & \cellcolor{tabsecond}{0.9696} & \cellcolor{tabfirst}{0.0136}& 
 \cellcolor{tabfirst}{26.17} &
 \cellcolor{tabfirst}{25.47}
 \\
 
\bottomrule
\end{tabular}
}
\vspace{-1mm}
\end{table*}

\vspace{-2pt}
\subsection{Experimental Setup}
\noindent{\bf Training dataset.}
We train on Flare7K~\cite{dai2022flare7k}, following its established synthesis pipeline. Background images from Flickr24K \cite{zhang2018single} are composited with reflective and scattering flares. To simulate incomplete off-frame scenarios, we apply luminance masks to define realistic regions for outpainting.

\vspace{2pt}
\noindent{\bf Evaluation dataset.}
We evaluate our approach on Flare7K's test set, comprising 100 real and 100 synthetic images in two scenarios: (1) no light source and (2) incomplete light sources created by shifting the boundary outward by 15 pixels from scenario (1). This setup assesses our method's effectiveness in \modify{handling incomplete} illumination.



\vspace{2pt}
\noindent{\bf Baseline methods.}
We compare with state-of-the-art SIFR methods: Zhou et al.~\cite{zhou2023improving}, Flare7K++~\cite{dai2023flare7k++}, and MFDNet~\cite{jiang2024mfdnet} \footnote{The recent work Difflare~\cite{zhou2024difflare} does not provide publicly available implementations or pre-trained models, making its inclusion in our comprehensive evaluation infeasible.}. Additionally, our diffusion-based outpainting approach compares with state-of-the-art diffusion-based
inpainting and outpainting models, including SD-Inpainting~\cite{rombach2022high}, SDXL-Inpainting~\cite{podell2023sdxl}, and PowerPaint~\cite{zhuang2024task}.

\vspace{-2pt}
\subsection{Quantitative Evaluations}
\noindent{\bf Comparisons with Existing Methods.}
\cref{tab:off_the_shelf} compares our approach with two categories of baselines: state-of-the-art SIFR models and diffusion-based outpainting methods. \modify{Additionally, we report G-PSNR and S-PSNR~\cite{dai2023flare7k++}, which assess flare removal performance in the glare and streak regions, as also shown in~\cref{tab:off_the_shelf}.} Our approach significantly boosts performance, notably increasing PSNR from 26.29 dB to 28.41 dB with Flare7K++~\cite{dai2023flare7k++} on real images without light sources. It also outperforms existing diffusion-based methods on both real and synthetic datasets, effectively addressing cases with incomplete illumination.

\begin{table}[t]
\caption{
\textbf{Quantitative evaluation of our proposed light source prediction method.}
We compare our multitask regression module against a baseline UNet-based approach using mIoU scores. These results demonstrate the effectiveness and reliability of our regression-based strategy.
}
\label{tab:lightsource}
\vspace{-3mm}
\centering
\footnotesize
\begin{tabular}{lcc}
\toprule
    Method & Flare7K Real & Flare7K Synthetic \\
    \midrule
    UNet  & 0.6216  & 0.6563  \\
    Ours  & \textbf{0.6310}  & \textbf{0.6619}  \\
    \bottomrule
\end{tabular}
\vspace{-1mm}
\end{table}

\vspace{2pt}
\noindent{\bf Evaluation of multitask regression-based light source prediction.}
\cref{tab:lightsource} compares our multitask regression-based module with a baseline UNet approach using mIoU. Our method achieves superior mIoU scores of 0.6310 (real) and 0.6619 (synthetic), outperforming the baseline's 0.6216 and 0.6563, respectively.

\begin{figure*}[t]
    \centering
    \includegraphics[width=\linewidth]{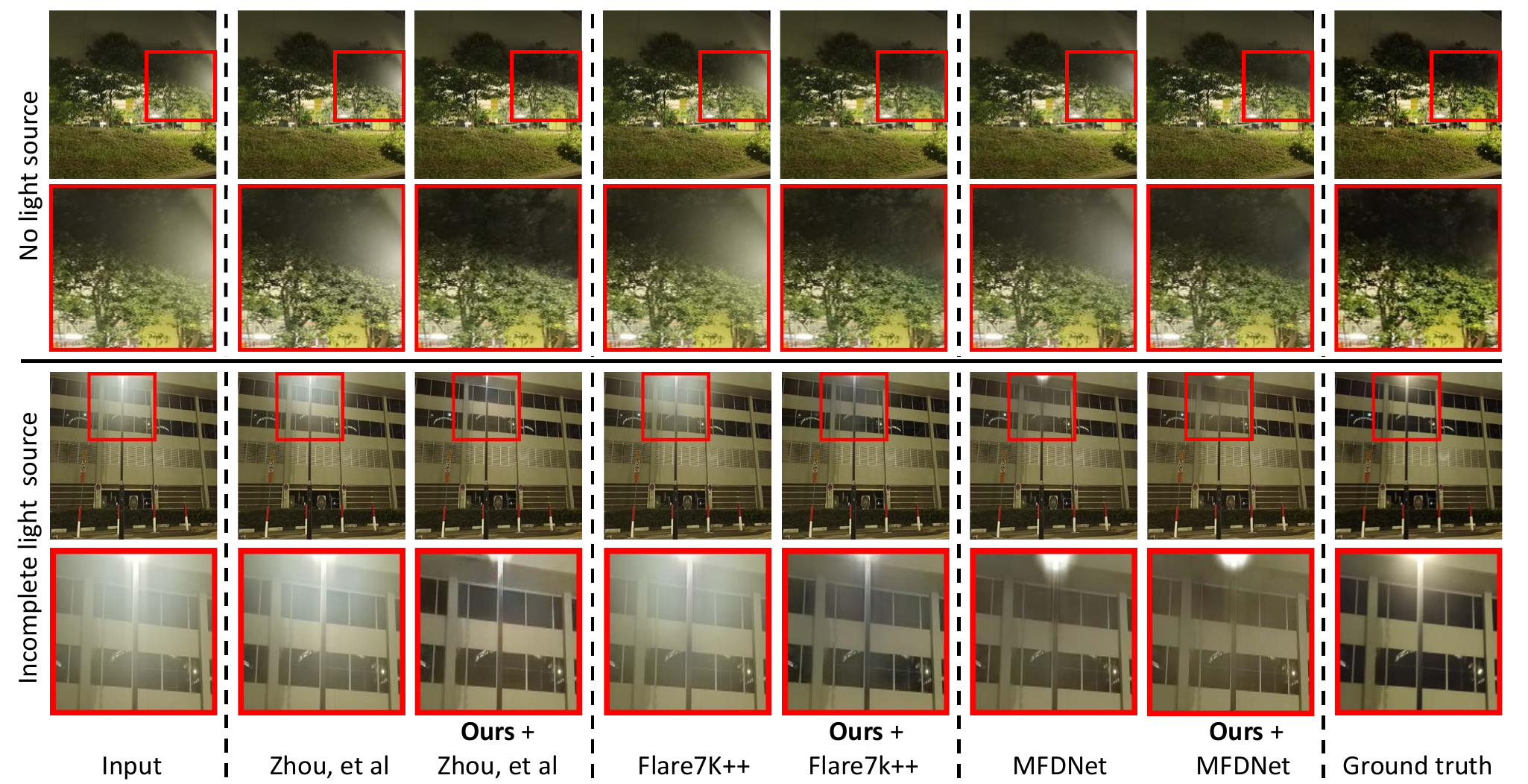}
\vspace{-7mm}
    \caption{
    \textbf{Qualitative comparison of lens flare removal results.} We compare state-of-the-art SIFR methods (Zhou et al.\cite{zhou2023improving}, Flare7K++\cite{dai2023flare7k++}, and MFDNet~\cite{jiang2024mfdnet}) alone and combined with our proposed method in two challenging scenarios: \modify{(\emph{top}) no visible light sources and (\emph{bottom}) incomplete light sources}. Integrating our outpainting method (``\textbf{Ours} +") significantly improves flare removal quality, producing results closer to ground truth.
    }
    \label{fig:qualitative_comparisons}
    \vspace{-3mm}
\end{figure*}

\vspace{-2pt}
\subsection{Qualitative Evaluations}
\noindent{\bf Qualitative impact of off-frame light source completion.}
\cref{fig:qualitative_comparisons} compares the qualitative results of existing SIFR models with and without our method. Without outpainting, these models produce noticeable residual flares and lower realism. Integrating our diffusion-based preprocessing significantly improves off-frame context, enabling more effective flare removal and results closely resembling ground truth.

\begin{figure}[t]
    \centering
    \includegraphics[width=\columnwidth]{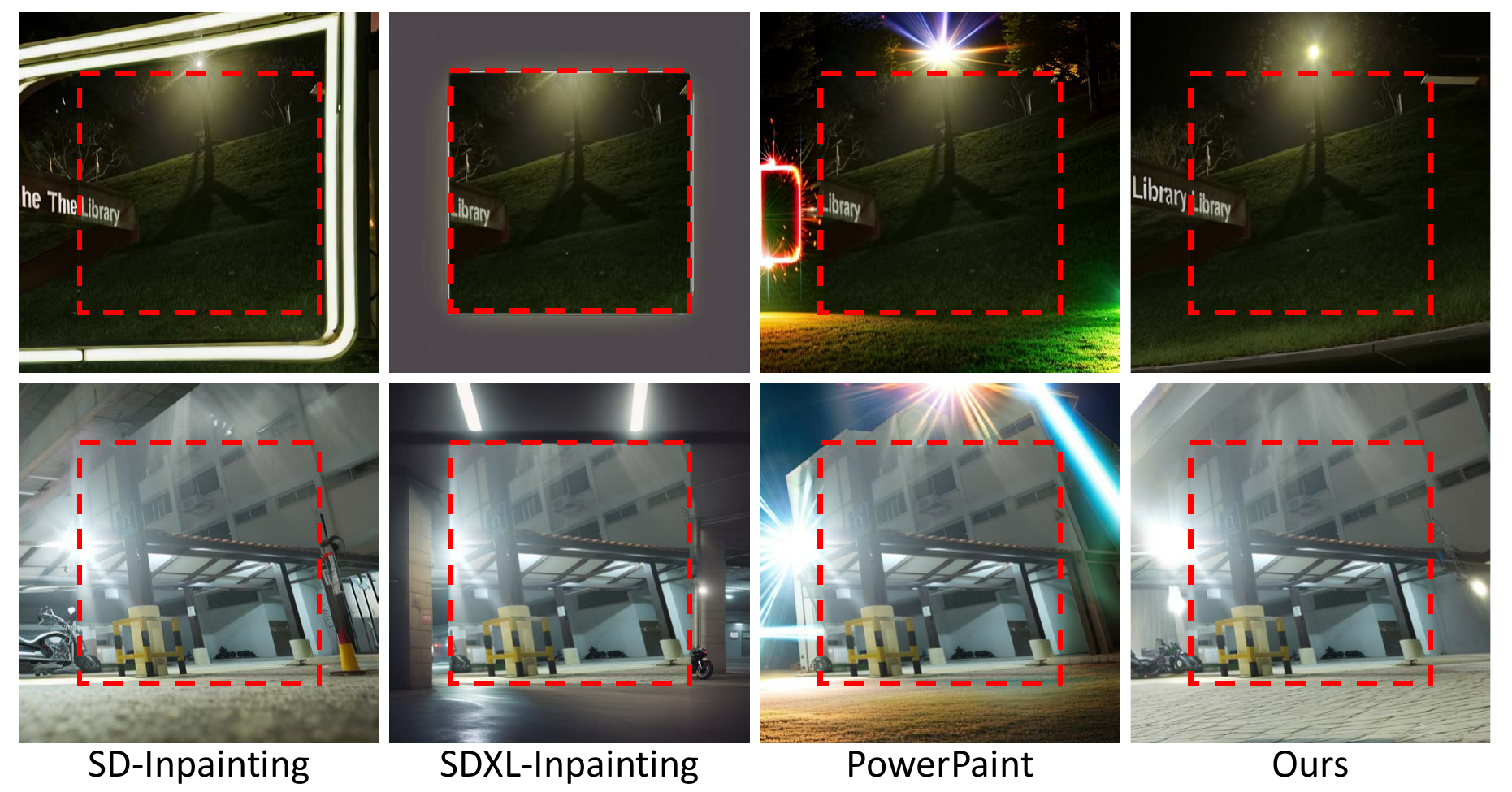}
\vspace{-7mm}
    \caption{
    \textbf{Qualitative comparison of our outpainting results.} 
    We qualitatively compare our method with SD-Inpainting~\cite{rombach2022high}, SDXL-Inpainting~\cite{podell2023sdxl}, and PowerPaint~\cite{zhuang2024task}. Our method produces more realistic outpainting results, accurately capturing flare artifacts and aligning closely with real-world scenes.
    }
    \label{fig:qualitative_comparisons_outpaint}
    \vspace{-3mm}
\end{figure}

\vspace{2pt}
\noindent{\bf Comparison with standard diffusion-based outpainting methods.}
We compare our outpainting results qualitatively against other diffusion-based methods in~\cref{fig:qualitative_comparisons_outpaint}. Standard diffusion methods often generate unrealistic off-frame content due to the lack of explicit conditioning on light sources and flare distributions. In contrast, LightsOut, with the multitask regression module and LoRA-fine-tuned diffusion model, ensures coherent outpainting with accurate flare artifacts and seamless illumination, closely matching the original scene and surpassing baseline methods.

\begin{figure*}[t]
    \centering
    \includegraphics[width=\textwidth]{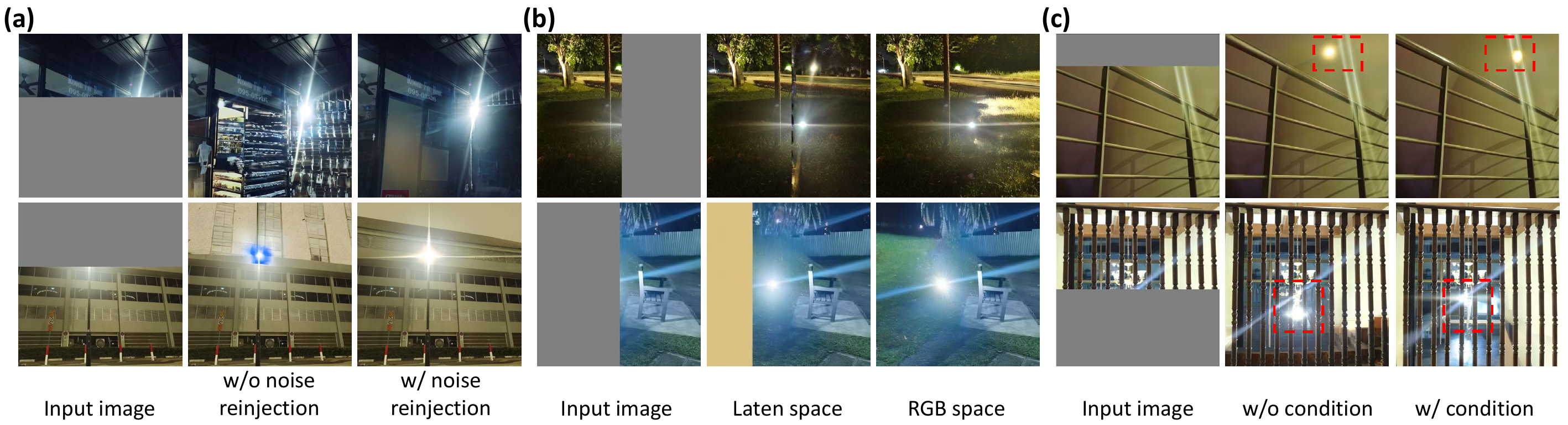}
\vspace{-7mm}
    \caption{\textbf{Ablation studies.} We ablate three components on the light source outpainting results:(a) Incorporating noise reinjection significantly enhances the diffusion model’s capability to produce realistic, seamless, and visually coherent outpainted regions; (b) Latent space blending tends to produce inconsistent illumination and noticeable artifacts, while blending in RGB space yields results with smoother transitions and improved alignment with real-world intensity distributions. (c) Integrating the proposed conditioning module significantly improves the accuracy and realism of the generated off-frame light sources and flare patterns.}
    \label{fig:ablation_3}
\vspace{-3mm}
\end{figure*}

\vspace{-2pt}
\subsection{Ablation Study}
\modify{
\noindent{\bf Effectiveness of the SIFR model trained for the off-frame setting.}
We retrained the original SIFR model using both incomplete and no-light-source data, while keeping all training hyperparameters identical to the original setup. As shown in \cref{tab:retrained}, despite being specifically designed and trained for the off-frame scenario, it still lags behind our method in performance. This confirms the effectiveness of our method in handling challenging cases where the light source lies outside the image frame.
}

\begin{table}[t]
\centering
\caption{
\modify{
\textbf{Ablation study on the SIFR model trained for the off-frame setting.} Although the baseline is explicitly designed and trained for the off-frame scenario, it still underperforms compared to our approach. $^{*}$ indicates that the Flare7k++ baseline was retrained using both incomplete and no-light-source data.
}
}
\label{tab:retrained}
\vspace{-3mm}
\resizebox{\columnwidth}{!}{%
\begin{tabular}{lccc ccc}
\toprule
\multirow{2}{*}{Method} & \multicolumn{3}{c}{Flare7k Real (no light source)} & \multicolumn{3}{c}{Flare7k Real (incomplete light source)} \\ \cmidrule(lr){2-4} \cmidrule(lr){5-7}
 &  PSNR $\uparrow$ & SSIM $\uparrow$ & LPIPS $\downarrow$ & PSNR $\uparrow$ & SSIM $\uparrow$ & LPIPS $\downarrow$\\
\midrule
Flare7k++$^{*}$ & 27.03 & 0.8679 & 0.0467 & 26.18 & 0.8650 & 0.0500 \\
Ours & \textbf{28.41} & \textbf{0.8956} & \textbf{0.0397} & \textbf{28.15} & \textbf{0.8957} & \textbf{0.0409} \\
\bottomrule
\end{tabular}
}
\vspace{-1mm}
\end{table}


\begin{table}[t]
\caption{
\textbf{Ablation study on noise reinjection strategy.}
We quantitatively evaluate the impact of noise reinjection during diffusion-based outpainting compared to a baseline without it. Results confirm the effectiveness of noise reinjection in enhancing flare removal performance.
}
\label{tab:ablation_noise}
\vspace{-3mm}
\centering
\resizebox{\columnwidth}{!}{%
\begin{tabular}{cccc ccc}
\toprule
\multirow{2}{*}{Noise regret} & \multicolumn{3}{c}{Flare7k Real} & \multicolumn{3}{c}{Flare7k Synthetic} \\ \cmidrule(lr){2-4} \cmidrule(lr){5-7}
 &  PSNR $\uparrow$ & SSIM $\uparrow$ & LPIPS $\downarrow$ & PSNR $\uparrow$ & SSIM $\uparrow$ & LPIPS $\downarrow$\\
\midrule
\ding{55} & 28.28 & 0.8949 & 0.0412 & 33.55 & 0.9704 & 0.0130 \\
\ding{51} & \textbf{28.41} & \textbf{0.8956} & \textbf{0.0397} & \textbf{33.91} & \textbf{0.9719} & \textbf{0.0120} \\
\bottomrule
\end{tabular}
}
\vspace{-1mm}
\end{table}


\vspace{2pt}
\noindent{\bf Effectiveness of noise reinjection.}
We quantitatively and qualitatively evaluate the impact of noise reinjection during diffusion-based outpainting. \cref{tab:ablation_noise} shows that excluding noise reinjection notably reduces performance, especially LPIPS scores. Qualitative comparisons (\cref{fig:ablation_3}(a)) confirm that noise reinjection significantly enhances visual coherence and realism, justifying its necessity.

\begin{table}[t]
\caption{
\textbf{Ablation study comparing latent space blending versus RGB space blending.}
We conduct a quantitative evaluation of performance disparities between two distinct blending methodologies: latent space blending and RGB space blending. The results confirms that RGB space blending is more effectively boosting the SIFR baselines.
}
\label{tab:ablation_blend}
\vspace{-3mm}
\centering
\footnotesize
\resizebox{\columnwidth}{!}{%
\begin{tabular}{lcccccc}
\toprule
\multirow{2}{*}{Blending} & \multicolumn{3}{c}{Flare7k Real} & \multicolumn{3}{c}{Flare7k Synthetic}  \\ \cmidrule(lr){2-4} \cmidrule(lr){5-7}
 &  PSNR $\uparrow$ & SSIM $\uparrow$ & LPIPS $\downarrow$ & PSNR $\uparrow$ & SSIM $\uparrow$ & LPIPS $\downarrow$ \\
\midrule
Latent space & 26.91 & \textbf{0.8859} & 0.0434 & 24.13 & 0.7156 & 0.0926\\
RGB space & \textbf{27.09} & 0.8856 & \textbf{0.0424} & \textbf{31.55 }& \textbf{0.9526} & \textbf{0.0178}\\
\bottomrule
\end{tabular}
}
\vspace{-1mm}
\end{table}


\vspace{2pt}
\noindent{\bf RGB vs. Latent space blending.}
We analyze two blending strategies: latent space and RGB space. As shown in~\cref{tab:ablation_blend}, RGB blending consistently outperforms latent space blending. Qualitative comparisons in~\cref{fig:ablation_3}(b) confirm that RGB blending offers smoother transitions and enhanced visual coherence, making it our preferred choice.


\vspace{2pt}
\noindent{\bf Impact of light source condition module.}
\modify{We} validate the contribution of our light source condition module by comparing results with and without conditioning (\cref{fig:ablation_3}(c)). Without conditioning, the diffusion model struggles to localize off-frame sources. Incorporating our module consistently yields more accurate and realistic predictions.

\begin{table}[t]
\caption{
\textbf{Ablation on optimization strategies for light source prediction.}
We compare three strategies: differentiable rendering~\cite{10.1145/3414685.3417871}, standard regression, and our multitask regression. Our multitask regression approach achieves superior mIoU, validating its simplicity and effectiveness for predicting off-frame light sources.
}
\label{tab:ablation_blend}
\vspace{-3mm}
\centering
\footnotesize
\begin{tabular}{lcc}
    \toprule
    Optimization Strategy & Flare7K Real & Flare7K Synthetic \\
    \midrule
    Differentiable rendering~\cite{10.1145/3414685.3417871} & 0.5212 & 0.5077 \\
    Regression with weighted loss & 0.6081 & 0.6577 \\
    Multitask regression (ours) & \textbf{0.6310}  & \textbf{0.6619}  \\
    \bottomrule
\end{tabular}
\vspace{-1mm}
\end{table}

\vspace{2pt}
\noindent{\bf Multitask regression for light source prediction.}
We evaluate our multitask regression against alternatives, including differentiable rendering~\cite{10.1145/3414685.3417871} and weighted-sum loss formulations. \cref{tab:ablation_blend} shows that our method consistently achieves superior accuracy, validating its robustness and design effectiveness.

\begin{table}[t]
\centering
\caption{
\modify{
\textbf{Ablation on the performance gap between SD-Inpainting and ours.}
To assess the effectiveness of each component, we compare LoRA fine-tuning and our light source conditioning module. The results demonstrate that the combination of all components yields the best performance, indicating that each component is essential.
}
}
\label{tab:performance_gap}
\vspace{-3mm}
\resizebox{\columnwidth}{!}{%
\begin{tabular}{ccccccc}
\toprule
\multirow{2}{*}{LoRA fine-tuning} & \multirow{2}{*}{Condition Module} & \multicolumn{5}{c}{Flare7k Real}  \\ \cmidrule(lr){3-7} 
 & &  PSNR $\uparrow$ & SSIM $\uparrow$ & LPIPS $\downarrow$ & G-PSNR $\uparrow$ & S-PSNR $\uparrow$\\
\midrule
- & - & 26.82 & 0.8886 & 0.0483 & 24.59 & 26.33 \\
\ding{51} & - & 27.12 & 0.8926 & 0.0453 & 24.83 & 26.59 \\
- & \ding{51} & 27.06 & 0.8906 & 0.0456 & 24.84 & 26.36 \\
\ding{51} & \ding{51} & \textbf{27.43} & \textbf{0.8940} & \textbf{0.0451} & \textbf{25.21} & \textbf{26.79} \\
\bottomrule
\end{tabular}
}
\vspace{-1mm}
\end{table}

\vspace{2pt}
\noindent{\bf \modify{Performance gap between SD-Inpainting and proposed method.}}
\modify{To analyze the performance gap, we ablate two key components: (1) LoRA fine-tuning and (2) our light source condition module. As shown in~\cref{tab:performance_gap}, each yields a PSNR gain of 0.30/0.24 dB and an LPIPS drop of 0.0030/0.0027. Combining both achieves the best overall performance, confirming their complementary contributions.}
\vspace{-2pt}
\section{Conclusion}
\label{sec:conclusion}
\vspace{-2pt}
LightsOut addresses the limitation of SIFR models caused by incomplete light source context. Our diffusion-based outpainting and conditioning modules effectively reconstruct off-frame illumination, significantly enhancing the flare removal performance of existing methods.

\vspace{2pt}
\noindent{\bf Limitations.} 
The added outpainting stages introduce computational overhead. Future work could explore end-to-end optimization strategies to reduce this overhead.

\newpage
\paragraph{Acknowledgements.}
This research was funded by the National Science and Technology Council, Taiwan, under Grants NSTC 112-2222-E-A49-004-MY2 and 113-2628-E-A49-023-. The authors are grateful to Google, NVIDIA, and MediaTek Inc. for their generous donations. Yu-Lun Liu acknowledges the Yushan Young Fellow Program by the MOE in Taiwan.

{\small
\bibliographystyle{ieeenat_fullname}
\bibliography{11_references}

\begin{thebibliography}{90}
\providecommand{\natexlab}[1]{#1}
\providecommand{\url}[1]{\texttt{#1}}
\expandafter\ifx\csname urlstyle\endcsname\relax
  \providecommand{\doi}[1]{doi: #1}\else
  \providecommand{\doi}{doi: \begingroup \urlstyle{rm}\Url}\fi

\bibitem[Asha et~al.(2019)Asha, Bhat, Nayak, and Bhat]{asha2019auto}
CS Asha, Sooraj~Kumar Bhat, Deepa Nayak, and Chaithra Bhat.
\newblock Auto removal of bright spot from images captured against flashing light source.
\newblock In \emph{2019 IEEE International Conference on Distributed Computing, VLSI, Electrical Circuits and Robotics}, 2019.

\bibitem[Ballester et~al.(2001)Ballester, Bertalmio, Caselles, Sapiro, and Verdera]{ballester2001filling}
Coloma Ballester, Marcelo Bertalmio, Vicent Caselles, Guillermo Sapiro, and Joan Verdera.
\newblock Filling-in by joint interpolation of vector fields and gray levels.
\newblock \emph{IEEE TIP}, 2001.

\bibitem[Bertalmio et~al.(2000)Bertalmio, Sapiro, Caselles, and Ballester]{Bertalmio2000image}
Marcelo Bertalmio, Guillermo Sapiro, Vincent Caselles, and Coloma Ballester.
\newblock Image inpainting.
\newblock In \emph{Proceedings of the 27th annual conference on Computer graphics and interactive techniques}, 2000.

\bibitem[Bertalmio et~al.(2003)Bertalmio, Vese, Sapiro, and Osher]{bertalmio2003simultaneous}
Marcelo Bertalmio, Luminita Vese, Guillermo Sapiro, and Stanley Osher.
\newblock Simultaneous structure and texture image inpainting.
\newblock \emph{IEEE TIP}, 2003.

\bibitem[Brooks et~al.(2023)Brooks, Holynski, and Efros]{brooks2023instructpix2pix}
Tim Brooks, Aleksander Holynski, and Alexei~A Efros.
\newblock Instructpix2pix: Learning to follow image editing instructions.
\newblock In \emph{CVPR}, 2023.

\bibitem[Carion et~al.(2020)Carion, Massa, Synnaeve, Usunier, Kirillov, and Zagoruyko]{carion2020end}
Nicolas Carion, Francisco Massa, Gabriel Synnaeve, Nicolas Usunier, Alexander Kirillov, and Sergey Zagoruyko.
\newblock End-to-end object detection with transformers.
\newblock In \emph{ECCV}, 2020.

\bibitem[Chabert(2014)]{automated_removal}
Floris Chabert.
\newblock Automated lens flare removal, 2014.
\newblock Technical Report, Department of Electrical Engineering, Stanford University.

\bibitem[Chao et~al.(2022)Chao, Sun, Cheng, Lo, Chang, Liu, Chang, Chen, and Lee]{chao2022denoising}
Chen-Hao Chao, Wei-Fang Sun, Bo-Wun Cheng, Yi-Chen Lo, Chia-Che Chang, Yu-Lun Liu, Yu-Lin Chang, Chia-Ping Chen, and Chun-Yi Lee.
\newblock Denoising likelihood score matching for conditional score-based data generation.
\newblock \emph{arXiv preprint arXiv:2203.14206}, 2022.

\bibitem[Chen et~al.(2024)Chen, Chan, Shiu, Yen, Yeh, and Liu]{chen2024narcan}
Ting-Hsuan Chen, Jie~Wen Chan, Hau-Shiang Shiu, Shih-Han Yen, Changhan Yeh, and Yu-Lun Liu.
\newblock Narcan: Natural refined canonical image with integration of diffusion prior for video editing.
\newblock \emph{NeurIPS}, 2024.

\bibitem[Choi et~al.(2021)Choi, Kim, Jeong, Gwon, and Yoon]{ilvr}
Jooyoung Choi, Sungwon Kim, Yonghyun Jeong, Youngjune Gwon, and Sungroh Yoon.
\newblock Ilvr: Conditioning method for denoising diffusion probabilistic models.
\newblock \emph{arXiv preprint arXiv:2108.02938}, 2021.

\bibitem[Dai et~al.(2022)Dai, Li, Zhou, Feng, and Loy]{dai2022flare7k}
Yuekun Dai, Chongyi Li, Shangchen Zhou, Ruicheng Feng, and Chen~Change Loy.
\newblock Flare7k: A phenomenological nighttime flare removal dataset.
\newblock \emph{NeurIPS}, 2022.

\bibitem[Dai et~al.(2023)Dai, Luo, Zhou, Li, and Loy]{dai2023nighttime}
Yuekun Dai, Yihang Luo, Shangchen Zhou, Chongyi Li, and Chen~Change Loy.
\newblock Nighttime smartphone reflective flare removal using optical center symmetry prior.
\newblock In \emph{CVPR}, 2023.

\bibitem[Dai et~al.(2024)Dai, Li, Zhou, Feng, Luo, and Loy]{dai2023flare7k++}
Yuekun Dai, Chongyi Li, Shangchen Zhou, Ruicheng Feng, Yihang Luo, and Chen~Change Loy.
\newblock Flare7k++: Mixing synthetic and real datasets for nighttime flare removal and beyond.
\newblock \emph{IEEE TPAMI}, 2024.

\bibitem[Gal et~al.(2022)Gal, Alaluf, Atzmon, Patashnik, Bermano, Chechik, and Cohen-Or]{gal2022image}
Rinon Gal, Yuval Alaluf, Yuval Atzmon, Or Patashnik, Amit~H Bermano, Gal Chechik, and Daniel Cohen-Or.
\newblock An image is worth one word: Personalizing text-to-image generation using textual inversion.
\newblock \emph{arXiv preprint arXiv:2208.01618}, 2022.

\bibitem[Goodfellow et~al.(2014)Goodfellow, Pouget-Abadie, Mirza, Xu, Warde-Farley, Ozair, Courville, and Bengio]{gan}
Ian Goodfellow, Jean Pouget-Abadie, Mehdi Mirza, Bing Xu, David Warde-Farley, Sherjil Ozair, Aaron Courville, and Yoshua Bengio.
\newblock Generative adversarial nets.
\newblock \emph{NeurIPS}, 2014.

\bibitem[Guo et~al.(2020)Guo, Liu, Zhao, Cheng, Song, Gu, Zheng, and Zheng]{spiral}
Dongsheng Guo, Hongzhi Liu, Haoru Zhao, Yunhao Cheng, Qingwei Song, Zhaorui Gu, Haiyong Zheng, and Bing Zheng.
\newblock Spiral generative network for image extrapolation.
\newblock In \emph{ECCV}, 2020.

\bibitem[Guo et~al.(2021)Guo, Yang, and Huang]{guo2021image}
Xiefan Guo, Hongyu Yang, and Di Huang.
\newblock Image inpainting via conditional texture and structure dual generation.
\newblock In \emph{ICCV}, 2021.

\bibitem[Han et~al.(2023)Han, Li, Zhang, Milanfar, Metaxas, and Yang]{han2023svdiff}
Ligong Han, Yinxiao Li, Han Zhang, Peyman Milanfar, Dimitris Metaxas, and Feng Yang.
\newblock Svdiff: Compact parameter space for diffusion fine-tuning.
\newblock In \emph{ICCV}, 2023.

\bibitem[Hays and Efros(2007)]{hays2007scene}
James Hays and Alexei~A Efros.
\newblock Scene completion using millions of photographs.
\newblock \emph{ACM TOG}, 2007.

\bibitem[Ho et~al.(2020)Ho, Jain, and Abbeel]{ddpm}
Jonathan Ho, Ajay Jain, and Pieter Abbeel.
\newblock Denoising diffusion probabilistic models.
\newblock \emph{NeurIPS}, 2020.

\bibitem[Hochreiter and Schmidhuber(1997)]{lstm}
Sepp Hochreiter and J{\"u}rgen Schmidhuber.
\newblock Long short-term memory.
\newblock \emph{Neural Computation}, 1997.

\bibitem[{Holladay}(1926)]{1926JOSA...12..271H}
L.~L. {Holladay}.
\newblock {The Fundamentals of Glare and Visibility}.
\newblock \emph{Journal of the Optical Society of America (1917-1983)}, 1926.

\bibitem[Hong et~al.(2018)Hong, Yan, Huang, and Lee]{hong2018learning}
Seunghoon Hong, Xinchen Yan, Thomas Huang, and Honglak Lee.
\newblock Learning hierarchical semantic image manipulation through structured representations.
\newblock \emph{arXiv preprint arXiv:1808.07535}, 2018.

\bibitem[Hsiao et~al.(2024)Hsiao, Liu, Yang, Kuo, Jou, and Chen]{hsiao2024ref}
Chi-Wei Hsiao, Yu-Lun Liu, Cheng-Kun Yang, Sheng-Po Kuo, Kevin Jou, and Chia-Ping Chen.
\newblock Ref-ldm: A latent diffusion model for reference-based face image restoration.
\newblock \emph{NeurIPS}, 2024.

\bibitem[Hu et~al.(2022)Hu, Shen, Wallis, Allen-Zhu, Li, Wang, Wang, Chen, et~al.]{hu2022lora}
Edward~J Hu, Yelong Shen, Phillip Wallis, Zeyuan Allen-Zhu, Yuanzhi Li, Shean Wang, Lu Wang, Weizhu Chen, et~al.
\newblock Lora: Low-rank adaptation of large language models.
\newblock \emph{ICLR}, 2022.

\bibitem[Hullin et~al.(2011)Hullin, Eisemann, Seidel, and Lee]{10.1145/2010324.1965003}
Matthias Hullin, Elmar Eisemann, Hans-Peter Seidel, and Sungkil Lee.
\newblock Physically-based real-time lens flare rendering.
\newblock \emph{ACM TOG}, 2011.

\bibitem[Iizuka et~al.(2017)Iizuka, Simo-Serra, and Ishikawa]{10.1145/3072959.3073659}
Satoshi Iizuka, Edgar Simo-Serra, and Hiroshi Ishikawa.
\newblock Globally and locally consistent image completion.
\newblock \emph{ACM TOG}, 2017.

\bibitem[Jiang et~al.(2024)Jiang, Chen, Pun, Wang, and Feng]{jiang2024mfdnet}
Yiguo Jiang, Xuhang Chen, Chi-Man Pun, Shuqiang Wang, and Wei Feng.
\newblock Mfdnet: Multi-frequency deflare network for efficient nighttime flare removal.
\newblock \emph{The Visual Computer}, 2024.

\bibitem[Jocher and Qiu(2024)]{yolo11_ultralytics}
Glenn Jocher and Jing Qiu.
\newblock Ultralytics yolo11, 2024.

\bibitem[Kendall et~al.(2018)Kendall, Gal, and Cipolla]{kendall2018multi}
Alex Kendall, Yarin Gal, and Roberto Cipolla.
\newblock Multi-task learning using uncertainty to weigh losses for scene geometry and semantics.
\newblock In \emph{CVPR}, 2018.

\bibitem[Lee and Eisemann(2013)]{flare_simulation2}
Sungkil Lee and Elmar Eisemann.
\newblock Practical real-time lens-flare rendering.
\newblock \emph{Computer Graphics Forum}, 2013.

\bibitem[Li et~al.(2023)Li, Li, Savarese, and Hoi]{li2023blip}
Junnan Li, Dongxu Li, Silvio Savarese, and Steven Hoi.
\newblock Blip-2: Bootstrapping language-image pre-training with frozen image encoders and large language models.
\newblock In \emph{ICML}, 2023.

\bibitem[Li et~al.(2020)Li, Luk\'{a}\v{c}, Gharbi, and Ragan-Kelley]{10.1145/3414685.3417871}
Tzu-Mao Li, Michal Luk\'{a}\v{c}, Micha\"{e}l Gharbi, and Jonathan Ragan-Kelley.
\newblock Differentiable vector graphics rasterization for editing and learning.
\newblock \emph{ACM TOG}, 2020.

\bibitem[Li et~al.(2021)Li, Zhang, Liao, and Sander]{li2021let}
Xiaoyu Li, Bo Zhang, Jing Liao, and Pedro~V Sander.
\newblock Let's see clearly: Contaminant artifact removal for moving cameras.
\newblock In \emph{ICCV}, 2021.

\bibitem[Liebel and K{\"o}rner(2018)]{liebel2018auxiliary}
Lukas Liebel and Marco K{\"o}rner.
\newblock Auxiliary tasks in multi-task learning.
\newblock \emph{arXiv preprint arXiv:1805.06334}, 2018.

\bibitem[Lin et~al.(2021)Lin, Pagnucco, and Song]{edgeout}
Han Lin, Maurice Pagnucco, and Yang Song.
\newblock Edge guided progressively generative image outpainting.
\newblock In \emph{CVPR}, 2021.

\bibitem[Liu et~al.(2018)Liu, Reda, Shih, Wang, Tao, and Catanzaro]{liu2018image}
Guilin Liu, Fitsum~A Reda, Kevin~J Shih, Ting-Chun Wang, Andrew Tao, and Bryan Catanzaro.
\newblock Image inpainting for irregular holes using partial convolutions.
\newblock In \emph{ECCV}, 2018.

\bibitem[Liu et~al.(2025)Liu, Yang, Chen, Liu, and Lin]{liu2025corrfill}
Kuan-Hung Liu, Cheng-Kun Yang, Min-Hung Chen, Yu-Lun Liu, and Yen-Yu Lin.
\newblock Corrfill: Enhancing faithfulness in reference-based inpainting with correspondence guidance in diffusion models.
\newblock In \emph{WACV}, 2025.

\bibitem[Liu et~al.(2022)Liu, Li, Du, Torralba, and Tenenbaum]{compositional}
Nan Liu, Shuang Li, Yilun Du, Antonio Torralba, and Joshua~B Tenenbaum.
\newblock Compositional visual generation with composable diffusion models.
\newblock In \emph{ECCV}, 2022.

\bibitem[Lugmayr et~al.(2022)Lugmayr, Danelljan, Romero, Yu, Timofte, and Van~Gool]{lugmayr2022repaint}
Andreas Lugmayr, Martin Danelljan, Andres Romero, Fisher Yu, Radu Timofte, and Luc Van~Gool.
\newblock Repaint: Inpainting using denoising diffusion probabilistic models.
\newblock In \emph{CVPR}, 2022.

\bibitem[Meng et~al.(2021)Meng, He, Song, Song, Wu, Zhu, and Ermon]{sdedit}
Chenlin Meng, Yutong He, Yang Song, Jiaming Song, Jiajun Wu, Jun-Yan Zhu, and Stefano Ermon.
\newblock Sdedit: Guided image synthesis and editing with stochastic differential equations.
\newblock \emph{arXiv preprint arXiv:2108.01073}, 2021.

\bibitem[Mou et~al.(2024)Mou, Wang, Xie, Wu, Zhang, Qi, and Shan]{mou2024t2i}
Chong Mou, Xintao Wang, Liangbin Xie, Yanze Wu, Jian Zhang, Zhongang Qi, and Ying Shan.
\newblock T2i-adapter: Learning adapters to dig out more controllable ability for text-to-image diffusion models.
\newblock In \emph{AAAI}, 2024.

\bibitem[Nazeri et~al.(2019)Nazeri, Ng, Joseph, Qureshi, and Ebrahimi]{nazeri2019edgeconnect}
Kamyar Nazeri, Eric Ng, Tony Joseph, Faisal Qureshi, and Mehran Ebrahimi.
\newblock Edgeconnect: Structure guided image inpainting using edge prediction.
\newblock In \emph{ICCV Workshops}, 2019.

\bibitem[Nichol and Dhariwal(2021)]{improvedddpm}
Alex Nichol and Prafulla Dhariwal.
\newblock Improved denoising diffusion probabilistic models.
\newblock \emph{arXiv preprint arXiv:2102.09672}, 2021.

\bibitem[Nichol et~al.(2021)Nichol, Dhariwal, Ramesh, Shyam, Mishkin, McGrew, Sutskever, and Chen]{glide}
Alex Nichol, Prafulla Dhariwal, Aditya Ramesh, Pranav Shyam, Pamela Mishkin, Bob McGrew, Ilya Sutskever, and Mark Chen.
\newblock Glide: Towards photorealistic image generation and editing with text-guided diffusion models.
\newblock \emph{arXiv preprint arXiv:2112.10741}, 2021.

\bibitem[Ntavelis et~al.(2020)Ntavelis, Romero, Kastanis, Van~Gool, and Timofte]{ntavelis2020sesame}
Evangelos Ntavelis, Andr{\'e}s Romero, Iason Kastanis, Luc Van~Gool, and Radu Timofte.
\newblock Sesame: semantic editing of scenes by adding, manipulating or erasing objects.
\newblock In \emph{ECCV}, 2020.

\bibitem[Nussberger et~al.(2015)Nussberger, Grabner, and Gool]{aerial_tracking}
Andreas Nussberger, Helmut Grabner, and Luc~Van Gool.
\newblock Robust aerial object tracking in images with lens flare.
\newblock In \emph{ICRA}, 2015.

\bibitem[Pathak et~al.(2016)Pathak, Krahenbuhl, Donahue, Darrell, and Efros]{pathak2016context}
Deepak Pathak, Philipp Krahenbuhl, Jeff Donahue, Trevor Darrell, and Alexei~A Efros.
\newblock Context encoders: Feature learning by inpainting.
\newblock In \emph{CVPR}, 2016.

\bibitem[Podell et~al.(2023)Podell, English, Lacey, Blattmann, Dockhorn, M{\"u}ller, Penna, and Rombach]{podell2023sdxl}
Dustin Podell, Zion English, Kyle Lacey, Andreas Blattmann, Tim Dockhorn, Jonas M{\"u}ller, Joe Penna, and Robin Rombach.
\newblock Sdxl: Improving latent diffusion models for high-resolution image synthesis.
\newblock \emph{arXiv preprint arXiv:2307.01952}, 2023.

\bibitem[Qiao et~al.(2021)Qiao, Hancke, and Lau]{light_source}
Xiaotian Qiao, Gerhard~P. Hancke, and Rynson W.~H. Lau.
\newblock Light source guided single-image flare removal from unpaired data.
\newblock In \emph{ICCV}, 2021.

\bibitem[Qiu et~al.(2023)Qiu, Liu, Feng, Xue, Feng, Liu, Zhang, Weller, and Sch{\"o}lkopf]{qiu2023controlling}
Zeju Qiu, Weiyang Liu, Haiwen Feng, Yuxuan Xue, Yao Feng, Zhen Liu, Dan Zhang, Adrian Weller, and Bernhard Sch{\"o}lkopf.
\newblock Controlling text-to-image diffusion by orthogonal finetuning.
\newblock \emph{NeurIPS}, 2023.

\bibitem[Reinhard et~al.(2005)Reinhard, Ward, Pattanaik, and Debevec]{10.5555/1208706}
Erik Reinhard, Greg Ward, Sumanta Pattanaik, and Paul Debevec.
\newblock \emph{High Dynamic Range Imaging: Acquisition, Display, and Image-Based Lighting (The Morgan Kaufmann Series in Computer Graphics)}.
\newblock Morgan Kaufmann Publishers Inc., 2005.

\bibitem[Rombach et~al.(2022)Rombach, Blattmann, Lorenz, Esser, and Ommer]{rombach2022high}
Robin Rombach, Andreas Blattmann, Dominik Lorenz, Patrick Esser, and Bj{\"o}rn Ommer.
\newblock High-resolution image synthesis with latent diffusion models.
\newblock In \emph{CVPR}, 2022.

\bibitem[Ronneberger et~al.(2015)Ronneberger, Fischer, and Brox]{unet}
Olaf Ronneberger, Philipp Fischer, and Thomas Brox.
\newblock U-net: Convolutional networks for biomedical image segmentation.
\newblock In \emph{MICCAI}, 2015.

\bibitem[Ruiz et~al.(2023)Ruiz, Li, Jampani, Pritch, Rubinstein, and Aberman]{ruiz2023dreambooth}
Nataniel Ruiz, Yuanzhen Li, Varun Jampani, Yael Pritch, Michael Rubinstein, and Kfir Aberman.
\newblock Dreambooth: Fine tuning text-to-image diffusion models for subject-driven generation.
\newblock In \emph{CVPR}, 2023.

\bibitem[Saharia et~al.(2022{\natexlab{a}})Saharia, Chan, Chang, Lee, Ho, Salimans, Fleet, and Norouzi]{palette}
Chitwan Saharia, William Chan, Huiwen Chang, Chris Lee, Jonathan Ho, Tim Salimans, David Fleet, and Mohammad Norouzi.
\newblock Palette: Image-to-image diffusion models.
\newblock In \emph{ACM SIGGRAPH 2022 conference proceedings}, 2022{\natexlab{a}}.

\bibitem[Saharia et~al.(2022{\natexlab{b}})Saharia, Chan, Saxena, Li, Whang, Denton, Ghasemipour, Gontijo~Lopes, Karagol~Ayan, Salimans, et~al.]{saharia2022photorealistic}
Chitwan Saharia, William Chan, Saurabh Saxena, Lala Li, Jay Whang, Emily~L Denton, Kamyar Ghasemipour, Raphael Gontijo~Lopes, Burcu Karagol~Ayan, Tim Salimans, et~al.
\newblock Photorealistic text-to-image diffusion models with deep language understanding.
\newblock \emph{NeurIPS}, 2022{\natexlab{b}}.

\bibitem[Sohl-Dickstein et~al.(2015)Sohl-Dickstein, Weiss, Maheswaranathan, and Ganguli]{sohl2015deep}
Jascha Sohl-Dickstein, Eric Weiss, Niru Maheswaranathan, and Surya Ganguli.
\newblock Deep unsupervised learning using nonequilibrium thermodynamics.
\newblock In \emph{ICML}, 2015.

\bibitem[Song et~al.(2021)Song, Sohl-Dickstein, Kingma, Kumar, Ermon, and Poole]{song2021scorebased}
Yang Song, Jascha Sohl-Dickstein, Diederik~P Kingma, Abhishek Kumar, Stefano Ermon, and Ben Poole.
\newblock Score-based generative modeling through stochastic differential equations.
\newblock In \emph{ICLR}, 2021.

\bibitem[Talvala et~al.(2007)Talvala, Adams, Horowitz, and Levoy]{10.1145/1276377.1276424}
Eino-Ville Talvala, Andrew Adams, Mark Horowitz, and Marc Levoy.
\newblock Veiling glare in high dynamic range imaging.
\newblock \emph{ACM TOG}, 2007.

\bibitem[Tang et~al.(2024)Tang, Ruiz, Chu, Li, Holynski, Jacobs, Hariharan, Pritch, Wadhwa, Aberman, et~al.]{tang2024realfill}
Luming Tang, Nataniel Ruiz, Qinghao Chu, Yuanzhen Li, Aleksander Holynski, David~E Jacobs, Bharath Hariharan, Yael Pritch, Neal Wadhwa, Kfir Aberman, et~al.
\newblock Realfill: Reference-driven generation for authentic image completion.
\newblock \emph{ACM TOG}, 2024.

\bibitem[Vitoria and Ballester(2019)]{auto_removal2}
Patricia Vitoria and Coloma Ballester.
\newblock Automatic flare spot artifact detection and removal in photographs.
\newblock \emph{Journal of Mathematical Imaging and Vision}, 2019.

\bibitem[Voynov et~al.(2023)Voynov, Chu, Cohen-Or, and Aberman]{voynov2023p+}
Andrey Voynov, Qinghao Chu, Daniel Cohen-Or, and Kfir Aberman.
\newblock p+: Extended textual conditioning in text-to-image generation.
\newblock \emph{arXiv preprint arXiv:2303.09522}, 2023.

\bibitem[Wang et~al.(2024{\natexlab{a}})Wang, Wu, Huang, Shi, Shen, Song, Liu, and Li]{wang2024your}
Fu-Yun Wang, Xiaoshi Wu, Zhaoyang Huang, Xiaoyu Shi, Dazhong Shen, Guanglu Song, Yu Liu, and Hongsheng Li.
\newblock Be-your-outpainter: Mastering video outpainting through input-specific adaptation.
\newblock In \emph{ECCV}, 2024{\natexlab{a}}.

\bibitem[Wang et~al.(2022{\natexlab{a}})Wang, Bao, Zhou, Chen, Chen, Yuan, and Li]{sindiff}
Weilun Wang, Jianmin Bao, Wengang Zhou, Dongdong Chen, Dong Chen, Lu Yuan, and Houqiang Li.
\newblock Sindiffusion: Learning a diffusion model from a single natural image.
\newblock \emph{arXiv preprint arXiv:2211.12445}, 2022{\natexlab{a}}.

\bibitem[Wang et~al.(2019)Wang, Tao, Shen, and Jia]{srn}
Yi Wang, Xin Tao, Xiaoyong Shen, and Jiaya Jia.
\newblock Wide-context semantic image extrapolation.
\newblock In \emph{CVPR}, 2019.

\bibitem[Wang et~al.(2004)Wang, Bovik, Sheikh, and Simoncelli]{wang2004image}
Zhou Wang, Alan~C Bovik, Hamid~R Sheikh, and Eero~P Simoncelli.
\newblock Image quality assessment: from error visibility to structural similarity.
\newblock \emph{IEEE TIP}, 13\penalty0 (4):\penalty0 600--612, 2004.

\bibitem[Wang et~al.(2022{\natexlab{b}})Wang, Cun, Bao, Zhou, Liu, and Li]{uformer}
Zhendong Wang, Xiaodong Cun, Jianmin Bao, Wengang Zhou, Jianzhuang Liu, and Houqiang Li.
\newblock Uformer: A general u-shaped transformer for image restorationn.
\newblock In \emph{CVPR}, 2022{\natexlab{b}}.

\bibitem[Wang et~al.(2024{\natexlab{b}})Wang, Li, Wang, Liu, Gu, Chuang, and Satoh]{wang2024matting}
Zhixiang Wang, Baiang Li, Jian Wang, Yu-Lun Liu, Jinwei Gu, Yung-Yu Chuang, and Shin'Ichi Satoh.
\newblock Matting by generation.
\newblock In \emph{ACM SIGGRAPH 2024 Conference Papers}, 2024{\natexlab{b}}.

\bibitem[Wu et~al.(2025)Wu, Chen, Chen, Lee, Ke, Mu, Huang, Lin, Chen, Lin, et~al.]{wu2025aurafusion360}
Chung-Ho Wu, Yang-Jung Chen, Ying-Huan Chen, Jie-Ying Lee, Bo-Hsu Ke, Chun-Wei~Tuan Mu, Yi-Chuan Huang, Chin-Yang Lin, Min-Hung Chen, Yen-Yu Lin, et~al.
\newblock Aurafusion360: Augmented unseen region alignment for reference-based 360deg unbounded scene inpainting.
\newblock In \emph{CVPR}, 2025.

\bibitem[Wu et~al.(2021)Wu, He, Xue, Garg, Chen, Veeraraghavan, and Barron]{wu2021train}
Yicheng Wu, Qiurui He, Tianfan Xue, Rahul Garg, Jiawen Chen, Ashok Veeraraghavan, and Jonathan~T Barron.
\newblock How to train neural networks for flare removal.
\newblock In \emph{ICCV}, 2021.

\bibitem[Xiong et~al.(2019)Xiong, Yu, Lin, Yang, Lu, Barnes, and Luo]{xiong2019foreground}
Wei Xiong, Jiahui Yu, Zhe Lin, Jimei Yang, Xin Lu, Connelly Barnes, and Jiebo Luo.
\newblock Foreground-aware image inpainting.
\newblock In \emph{CVPR}, 2019.

\bibitem[Xu et~al.(2020)Xu, Liu, and Xiong]{xu2020e2i}
Shunxin Xu, Dong Liu, and Zhiwei Xiong.
\newblock E2i: Generative inpainting from edge to image.
\newblock \emph{IEEE Transactions on Circuits and Systems for Video Technology}, 2020.

\bibitem[Yang et~al.(2023)Yang, Gu, Zhang, Zhang, Chen, Sun, Chen, and Wen]{yang2023paint}
Binxin Yang, Shuyang Gu, Bo Zhang, Ting Zhang, Xuejin Chen, Xiaoyan Sun, Dong Chen, and Fang Wen.
\newblock Paint by example: Exemplar-based image editing with diffusion models.
\newblock In \emph{CVPR}, 2023.

\bibitem[Yang et~al.(2019)Yang, Dong, Liu, Yang, and Yan]{nsipo}
Zongxin Yang, Jian Dong, Ping Liu, Yi Yang, and Shuicheng Yan.
\newblock Very long natural scenery image prediction by outpainting.
\newblock In \emph{ICCV}, 2019.

\bibitem[Yao et~al.(2022)Yao, Gao, Yang, Sun, Zhang, and Huang]{queryqtr}
Kai Yao, Penglei Gao, Xi Yang, Jie Sun, Rui Zhang, and Kaizhu Huang.
\newblock Outpainting by queries.
\newblock In \emph{ECCV}, 2022.

\bibitem[Ye et~al.(2023)Ye, Zhang, Liu, Han, and Yang]{ye2023ip}
Hu Ye, Jun Zhang, Sibo Liu, Xiao Han, and Wei Yang.
\newblock Ip-adapter: Text compatible image prompt adapter for text-to-image diffusion models.
\newblock \emph{arXiv preprint arXiv:2308.06721}, 2023.

\bibitem[Yeh et~al.(2024)Yeh, Lin, Wang, Hsiao, Chen, Shiu, and Liu]{yeh2024diffir2vr}
Chang-Han Yeh, Chin-Yang Lin, Zhixiang Wang, Chi-Wei Hsiao, Ting-Hsuan Chen, Hau-Shiang Shiu, and Yu-Lun Liu.
\newblock Diffir2vr-zero: Zero-shot video restoration with diffusion-based image restoration models.
\newblock \emph{arXiv preprint arXiv:2407.01519}, 2024.

\bibitem[Yu and Koltun(2015)]{yu2015multi}
Fisher Yu and Vladlen Koltun.
\newblock Multi-scale context aggregation by dilated convolutions.
\newblock \emph{arXiv preprint arXiv:1511.07122}, 2015.

\bibitem[Yu et~al.(2018)Yu, Lin, Yang, Shen, Lu, and Huang]{yu2018generative}
Jiahui Yu, Zhe Lin, Jimei Yang, Xiaohui Shen, Xin Lu, and Thomas~S Huang.
\newblock Generative image inpainting with contextual attention.
\newblock In \emph{CVPR}, 2018.

\bibitem[Yu et~al.(2019)Yu, Lin, Yang, Shen, Lu, and Huang]{yu2019free}
Jiahui Yu, Zhe Lin, Jimei Yang, Xiaohui Shen, Xin Lu, and Thomas~S Huang.
\newblock Free-form image inpainting with gated convolution.
\newblock In \emph{ICCV}, 2019.

\bibitem[Zhang et~al.(2023)Zhang, Rao, and Agrawala]{zhang2023adding}
Lvmin Zhang, Anyi Rao, and Maneesh Agrawala.
\newblock Adding conditional control to text-to-image diffusion models.
\newblock In \emph{ICCV}, 2023.

\bibitem[Zhang et~al.(2018{\natexlab{a}})Zhang, Isola, Efros, Shechtman, and Wang]{zhang2018unreasonable}
Richard Zhang, Phillip Isola, Alexei~A Efros, Eli Shechtman, and Oliver Wang.
\newblock The unreasonable effectiveness of deep features as a perceptual metric.
\newblock In \emph{CVPR}, 2018{\natexlab{a}}.

\bibitem[Zhang et~al.(2018{\natexlab{b}})Zhang, Ng, and Chen]{zhang2018single}
Xuaner Zhang, Ren Ng, and Qifeng Chen.
\newblock Single image reflection separation with perceptual losses.
\newblock In \emph{CVPR}, 2018{\natexlab{b}}.

\bibitem[Zhou et~al.(2022)Zhou, Li, and Change~Loy]{zhou2022lednet}
Shangchen Zhou, Chongyi Li, and Chen Change~Loy.
\newblock {LEDNet}: Joint low-light enhancement and deblurring in the dark.
\newblock In \emph{ECCV}, 2022.

\bibitem[Zhou et~al.(2024)Zhou, Duan, and Yu]{zhou2024difflare}
Tianwen Zhou, Qihao Duan, and Zitong Yu.
\newblock Difflare: Removing image lens flare with latent diffusion model.
\newblock \emph{arXiv preprint arXiv:2407.14746}, 2024.

\bibitem[Zhou et~al.(2023)Zhou, Liang, Chen, Huang, Yang, and Li]{zhou2023improving}
Yuyan Zhou, Dong Liang, Songcan Chen, Sheng-Jun Huang, Shuo Yang, and Chongyi Li.
\newblock Improving lens flare removal with general-purpose pipeline and multiple light sources recovery.
\newblock In \emph{ICCV}, 2023.

\bibitem[Zhu et~al.(2017)Zhu, Park, Isola, and Efros]{CycleGAN}
Junyan Zhu, Taesung Park, Phillip Isola, and Alexei~A. Efros.
\newblock Unpaired image-to-image translation using cycle-consistent adversarial networks.
\newblock In \emph{ICCV}, 2017.

\bibitem[Zhu et~al.(2023)Zhu, Feng, Chen, Bao, Wang, Chen, Yuan, and Hua]{zhu2023designing}
Zixin Zhu, Xuelu Feng, Dongdong Chen, Jianmin Bao, Le Wang, Yinpeng Chen, Lu Yuan, and Gang Hua.
\newblock Designing a better asymmetric vqgan for stablediffusion.
\newblock \emph{arXiv preprint arXiv:2306.04632}, 2023.

\bibitem[Zhuang et~al.(2024)Zhuang, Zeng, Liu, Yuan, and Chen]{zhuang2024task}
Junhao Zhuang, Yanhong Zeng, Wenran Liu, Chun Yuan, and Kai Chen.
\newblock A task is worth one word: Learning with task prompts for high-quality versatile image inpainting.
\newblock In \emph{ECCV}, 2024.

\end{thebibliography}
}

\ifarxiv \clearpage \appendix \section{Appendix Section}
\label{sec:appendix_section}

\subsection{Implementation Details}
\vspace{3pt}
\noindent{\bf Dataset and Preprocessing.} We use the benchmark dataset Flare7k~\cite{dai2022flare7k} for both training and testing. Since the dataset was not originally designed for our tasks, we preprocess it to better suit our requirements. Specifically, to handle off-frame or incomplete light source images and define outpainted regions, we first generate YCbCr luminance masks and then apply an algorithm, formalized in \cref{algo:crop}, to identify the largest rectangular area in each image that excludes the light source. Once the bounding box is obtained, we crop the image on-the-fly during training and inference. The cropped region is then masked with a pixel value of 127, defining the area to be outpainted.

\begin{figure}[t]
    \centering
    \includegraphics[width=\columnwidth]{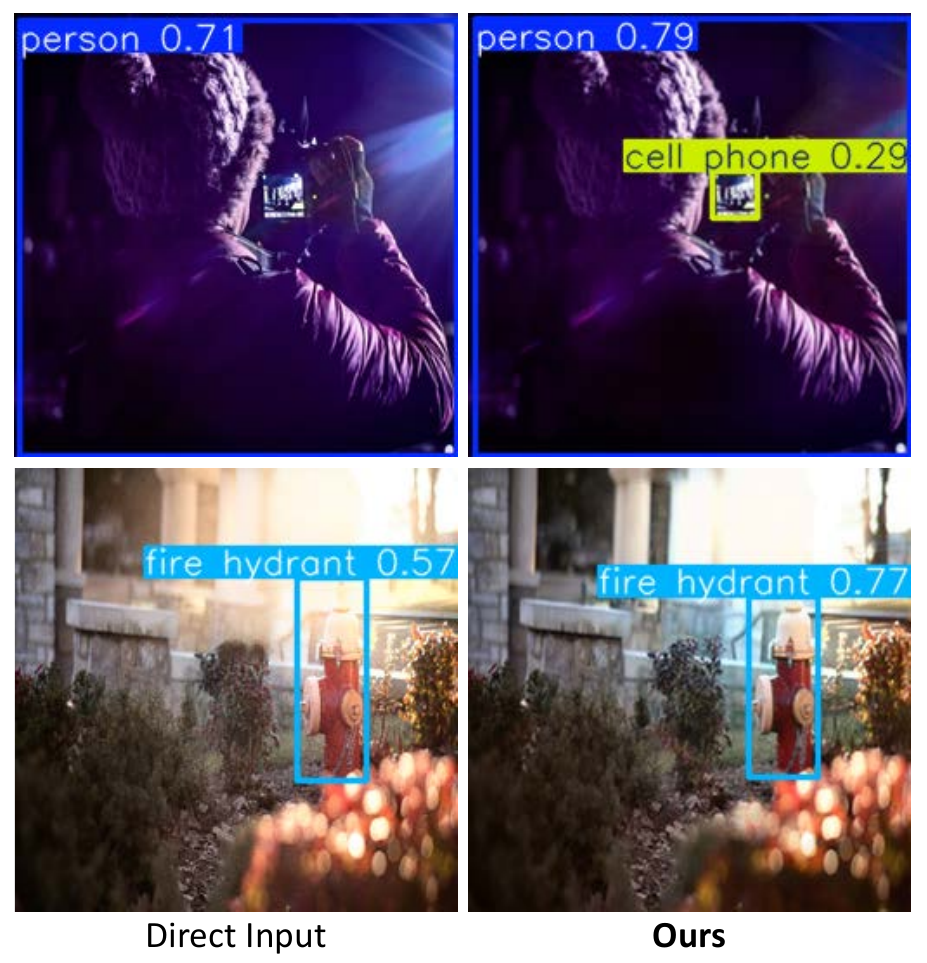}
    \caption{\textbf{Comparison of the downstream tasks.} The visual results indicate that LightsOut enhances performance on object detection tasks as well. Our approach not only boosts detection confidence scores but also enables the identification of objects previously undetectable due to flare artifacts.}
    \label{fig:downstream_task}
\end{figure}
\begin{algorithm}
\small
\caption{Cropping Algorithm}
\label{algo:crop}

\begin{algorithmic}[1]
\Function{ImageCrop}{\textit{image}}
    \Function{LargestRectangle}{\textit{heights}}
      \State \textit{heights}.\text{append}($0$)
      \State \textit{stack} $\gets [-1]$
      \State \textit{max\_area} $\gets 0$
      \State \textit{max\_bbox} $\gets (0, 0, 0, 0)$ \Comment{(area, left, right, height)}
      \For{$i \gets 0$ to $\text{len}(\textit{heights}) - 1$}
        \While{\textit{heights}[$i$] < \textit{heights}[\textit{stack}[{-}1]]}
          \State $h \gets \textit{heights}[\textit{stack}.\text{pop}()]$
          \State $w \gets i - \textit{stack}[{-}1] - 1$
          \State $\textit{area} \gets h \times w$
          \If{$\textit{area} > \textit{max\_area}$}
            \State $\textit{max\_area} \gets \textit{area}$
            \State $\textit{max\_bbox} \gets (\textit{area}, \textit{stack}[{-}1] + 1,\, i - 1,\, h)$
          \EndIf
        \EndWhile
        \State \textit{stack}.\text{append}($i$)
      \EndFor
      \State \Return \textit{max\_bbox}
    \EndFunction
    
  \State $\textit{max\_area} \gets 0$
  \State $\textit{max\_bbox} \gets [0, 0, 0, 0]$
  \State $\textit{heights} \gets \text{zeros\_like}(\textit{image.shape}[1])$

  \For{$\textit{row} \gets 0$ to $\textit{image.shape}[0] - 1$}
    \State $\textit{temp} \gets 1 - \textit{image}[\textit{row}]$
    \State $\textit{heights} \gets (\textit{heights} + \textit{temp}) \times \textit{temp}$
    \State $(\textit{area}, \textit{left}, \textit{right}, \textit{height}) \gets \text{LargestRectangle}(\textit{heights})$
    \If{$\textit{area} > \textit{max\_area}$}
      \State $\textit{max\_area} \gets \textit{area}$
      \State $\textit{max\_bbox} \gets [\textit{left},\; \textit{right},\; (\textit{row} - \textit{height} + 1),\; \textit{row}]$
    \EndIf
  \EndFor

  \State \Return $\textit{max\_bbox}$
\EndFunction

\end{algorithmic}
\end{algorithm}

\begin{figure*}[t]
    \centering
    \includegraphics[width=\textwidth]{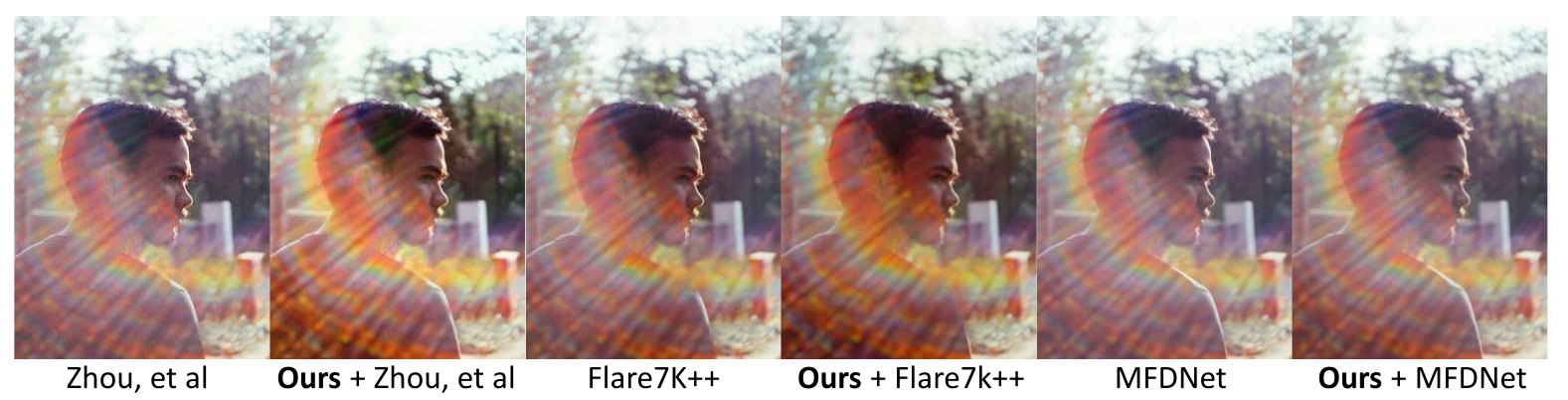}
    \caption{\textbf{Failure Cases.}}
    \label{fig:fail_case}
\end{figure*}

\vspace{3pt}
\noindent{\bf Training Details.} Our framework comprises three independently trained modules, all implemented on an NVIDIA RTX4090 GPU. The components are optimized independently, allowing each module to specialize in a distinct subtask and enabling them to collectively improve the system’s overall performance when integrated. The multitask regression module was trained with a learning rate of $1\times10^{-4}$, batch size of 32, for 100 epochs, and we set the number of predicted light sources $N$ to 4. The light source condition module was optimized using a learning rate of $1\times10^{-5}$ and a batch size of 8 for 20,000 steps. Finally, the Stable Diffusion inpainting network~\cite{rombach2022high} was fine-tuned using LoRA~\cite{hu2022lora} with a learning rate of $1\times10^{-4}$ and a batch size of 8 for 25,000 steps to achieve optimal performance while maintaining computational efficiency.

\vspace{3pt}
\noindent{\bf Inference Settings.} During outpainting process, we set the number of sampling steps to 50, the guidance scale to 7.0, and perform noise reinjection 4 times. Additionally, we utilize BLIP-2~\cite{li2023blip} to automatically generate captions, thereby minimizing human bias.

\vspace{3pt}
\noindent{\bf Evaluation metrics.}
We evaluate flare removal quality using PSNR, SSIM~\cite{wang2004image}, and LPIPS~\cite{zhang2018unreasonable}, and assess the accuracy of our light source prediction using mean Intersection over Union (mIoU).

\subsection{Downstream Tasks} Lens flare artifacts can negatively impact images in various computer vision tasks. To examine how flare removal affects object detection performance, we utilize the pre-trained YOLOv11~\cite{yolo11_ultralytics} detector to compare two scenarios: images directly processed by SIFR models, and images first enhanced by our proposed outpainting approach before being input to SIFR models. \cref{fig:downstream_task} demonstrates that our proposed approach yields improvements in detection accuracy, particularly for objects located in regions previously compromised by flare artifacts.

\subsection{In-the-Wild Images.} We present additional outpainting results on self-collected in-the-wild scenes in \cref{fig:wild_outpaint}, along with flare removal comparisons against baseline methods (Zhou et al.\cite{zhou2023improving}, Flare7K++\cite{dai2023flare7k++}, and MFDNet~\cite{jiang2024mfdnet}) in \cref{fig:wild_remove}. These results highlight our method's effectiveness in outpainting off-frame regions and improving the performance of existing SIFR models, even on challenging in-the-wild images.

\begin{figure}[t]
    \centering
    \includegraphics[width=\columnwidth]{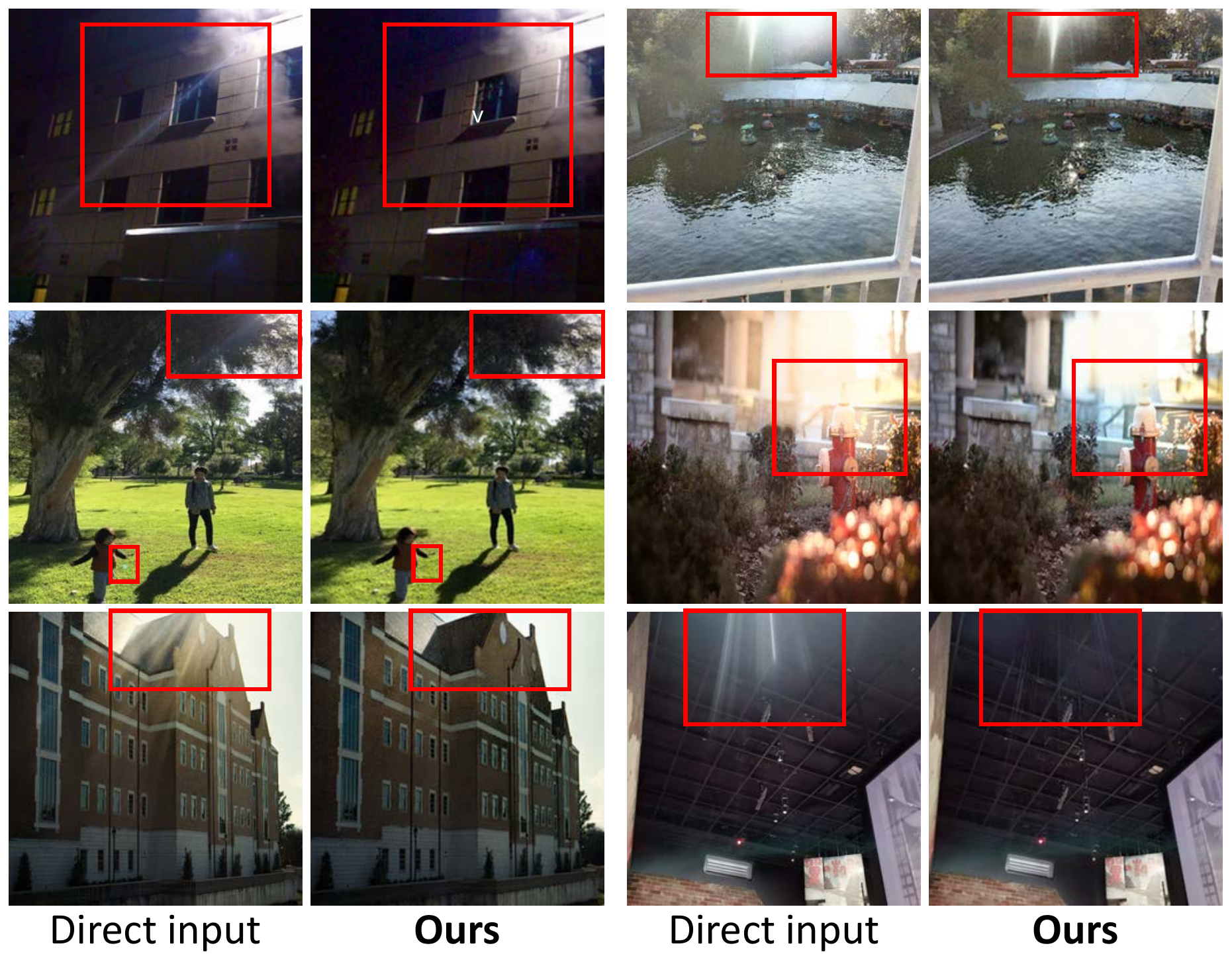}
    \caption{\textbf{Flare removal results for in-the-wild scens.} The red boxes indicate flare regions in the images. Our method effectively addresses off-frame light source scenes, which existing SIFR models fail to handle.}
    \label{fig:wild_remove}
\end{figure}

\begin{figure*}[t]
    \centering
    \includegraphics[width=\textwidth]{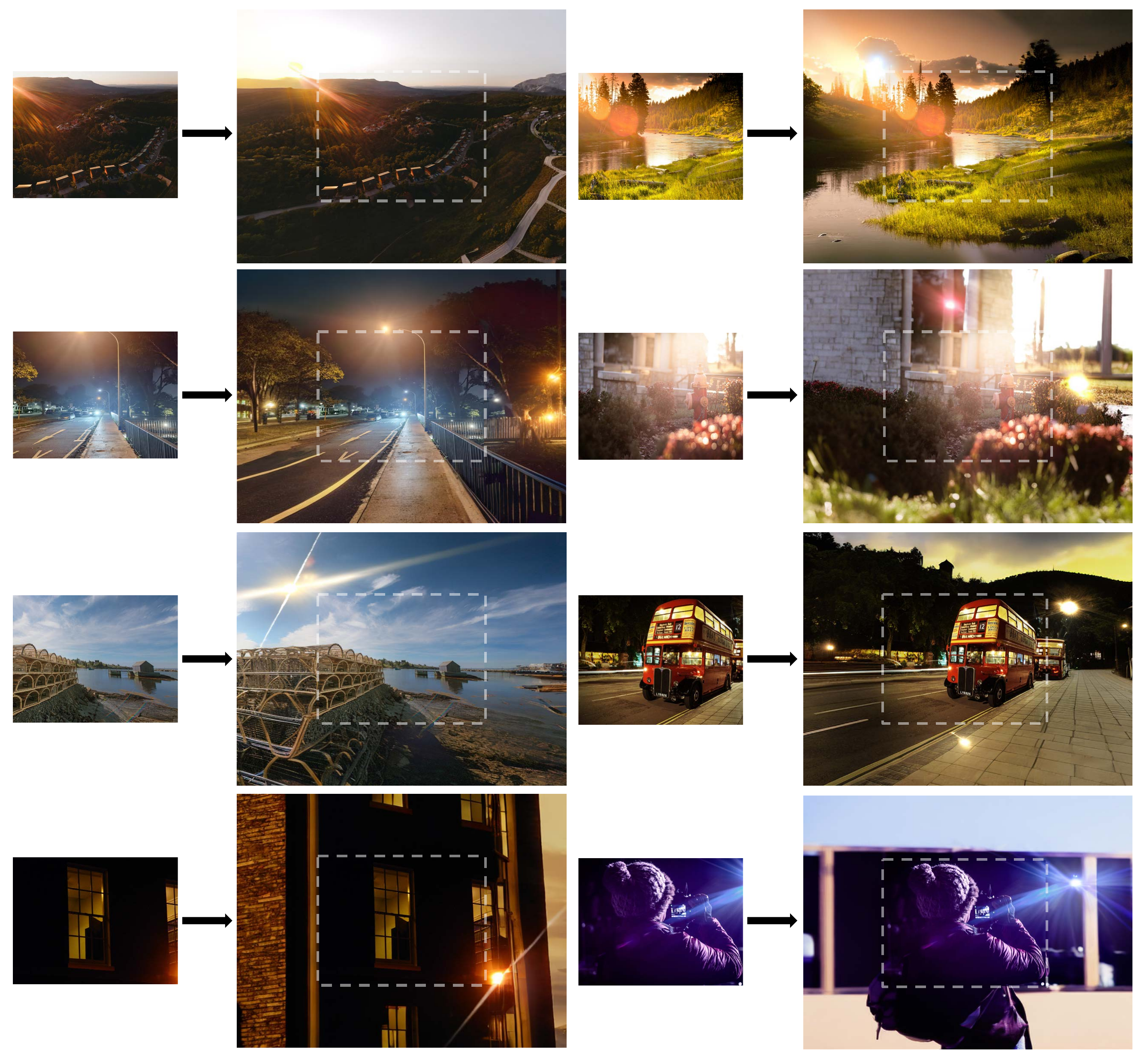}
    \caption{\textbf{Outpainting results for in-the-wild scens.}}
    \label{fig:wild_outpaint}
\end{figure*}

\subsection{Failure Cases} 
The main failure cases exhibit two characteristic features. First, when the overall image brightness is high, the brightness differential between the flare and other parts of the image becomes less pronounced. Second, when the flare occupies a relatively large proportion of the entire image. Both scenarios make it difficult to delineate the flare region precisely, even with the integration of our proposed method.

\subsection{Qualitative Comparisons of Light Source Mask Prediction.}
\cref{fig:qualitative_mask} compares the light source predictions from our multitask regression module with those generated by U-Net~\cite{unet}. The results demonstrate that our proposed module predicts the positions and radii of light sources more accurately, both in single and multiple light source scenarios.
\begin{figure*}[t]
    \centering
    \includegraphics[width=\textwidth]{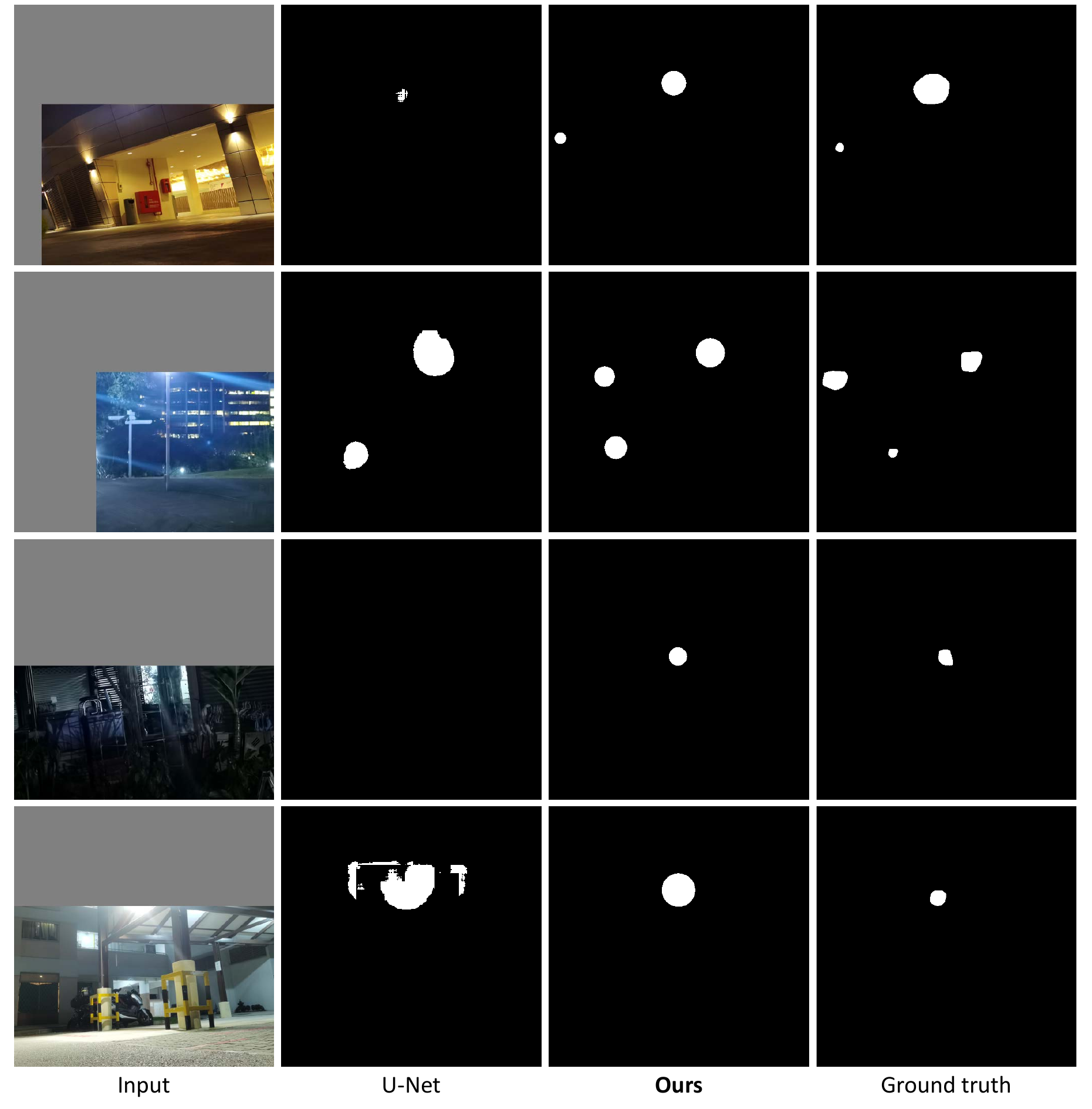}
    \caption{\textbf{Qualitative comparisons of light source mask prediction.} .}
    \label{fig:qualitative_mask}
\end{figure*}

\subsection{Additional Qualitative Comparisons}
We present extensive supplementary visual evidence to demonstrate the efficacy of our approach. Figures \cref{fig:additional_qualitative_3}, and \cref{fig:additional_qualitative_4} showcase additional flare removal results across diverse imaging conditions. Furthermore, we provide comparative analyses between our outpainting results and those produced by both baseline methods and state-of-the-art diffusion-based inpainting and outpainting techniques in \cref{fig:additional_qualitative_5}, \cref{fig:additional_qualitative_6}. These comprehensive visual comparisons substantiate the superior robustness and effectiveness of our proposed methodology across a wide spectrum of challenging scenarios.


\begin{figure*}[t]
    \centering
    \includegraphics[width=\textwidth]{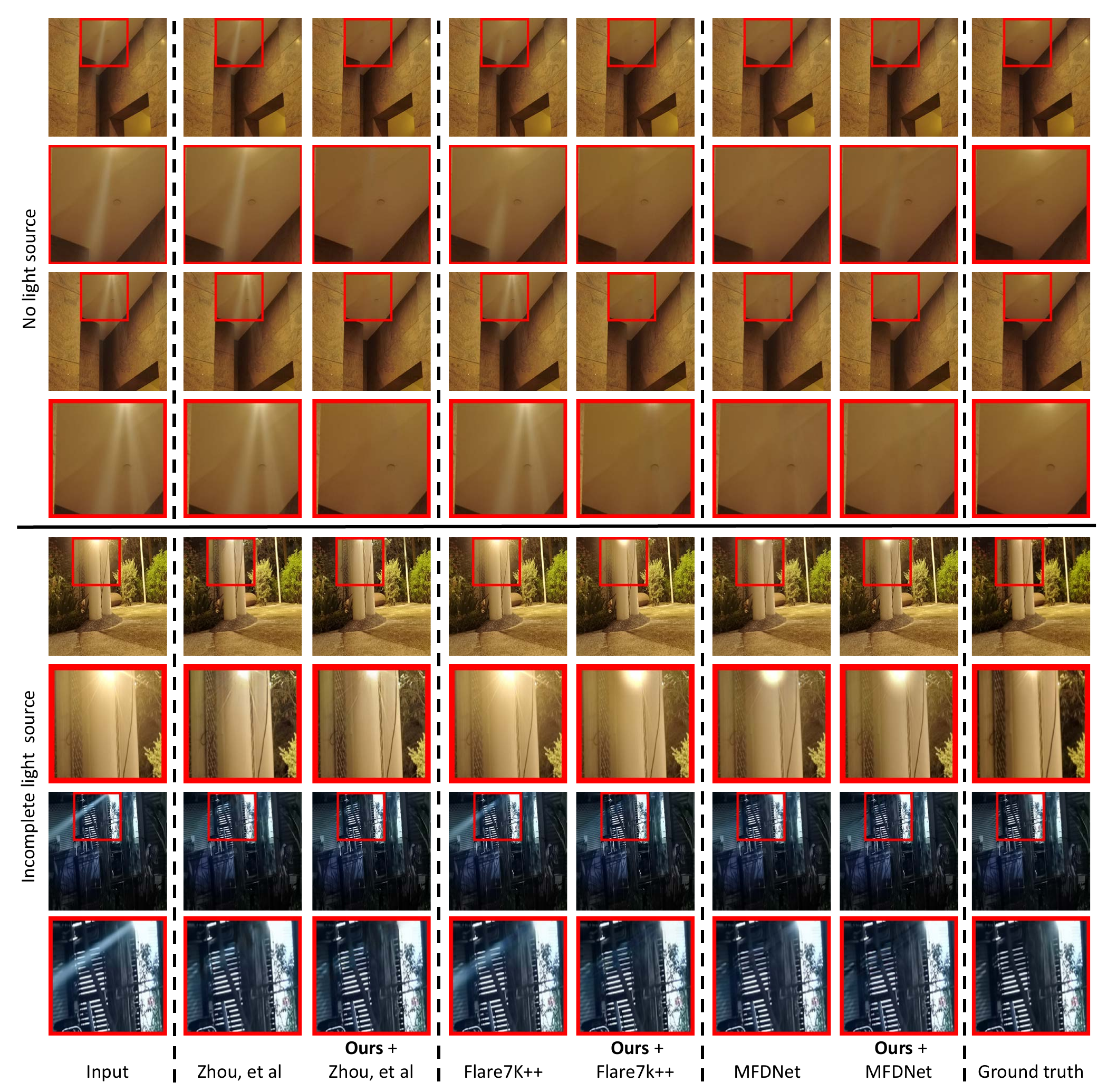}
    \caption{\textbf{Additional Qualitative Comparisons.} .}
    \label{fig:additional_qualitative_3}
\end{figure*}

\begin{figure*}[t]
    \centering
    \includegraphics[width=\textwidth]{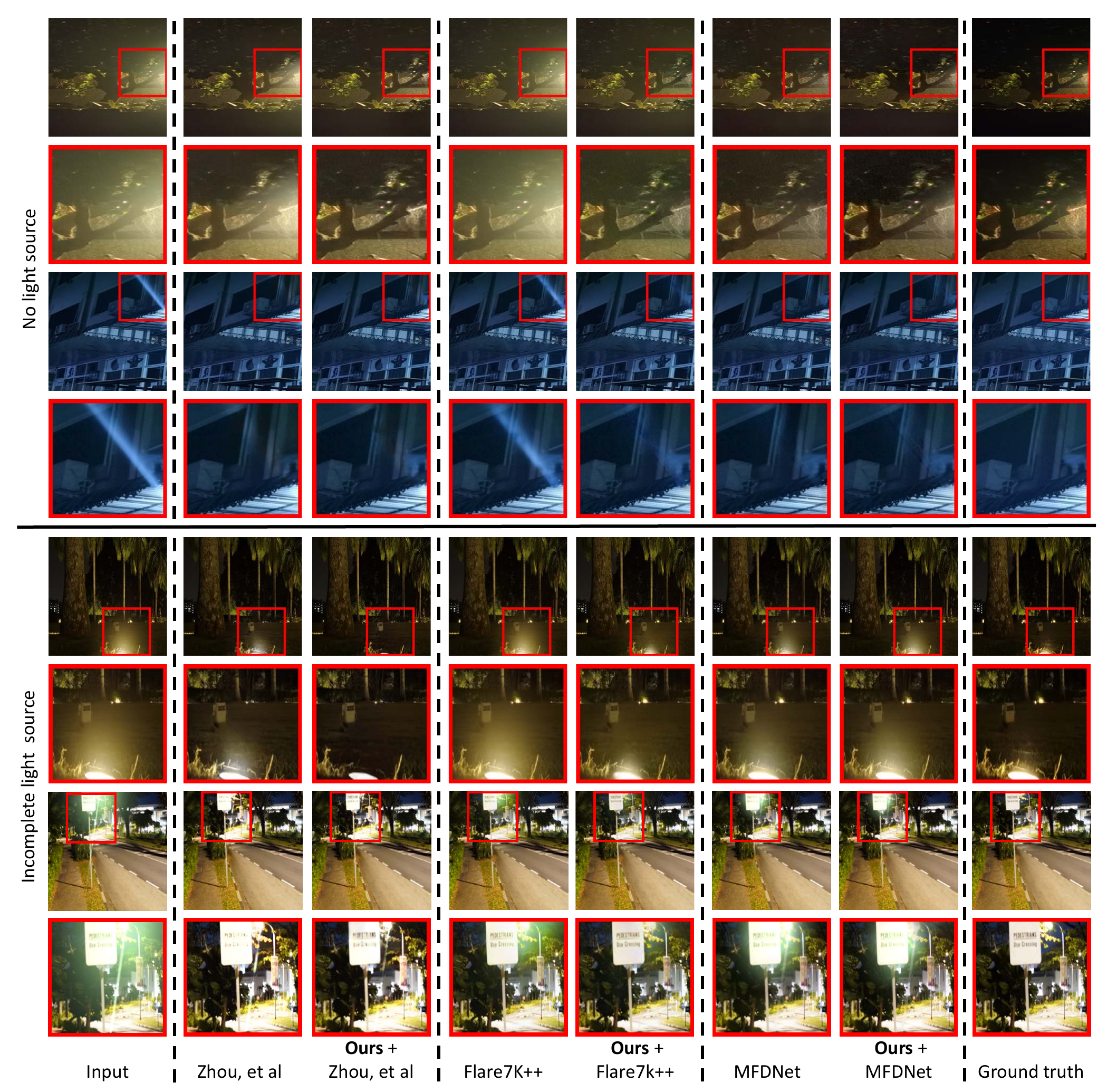}
    \caption{\textbf{Additional Qualitative Comparisons.} .}
    \label{fig:additional_qualitative_4}
\end{figure*}

\begin{figure*}[t]
    \centering
    \includegraphics[width=\textwidth]{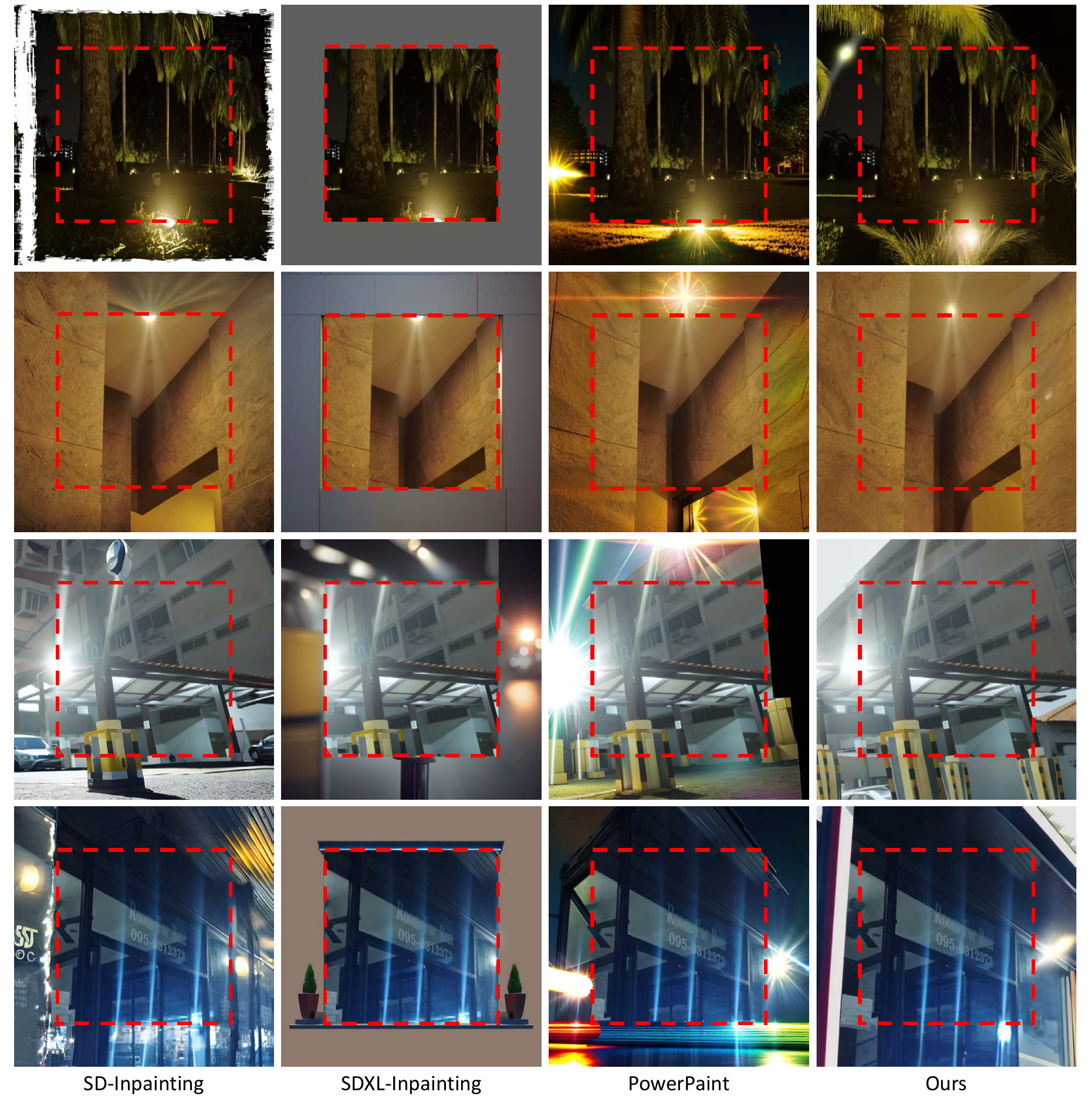}
    \caption{\textbf{Additional Qualitative Comparisons.} .}
    \label{fig:additional_qualitative_5}
\end{figure*}

\begin{figure*}[t]
    \centering
    \includegraphics[width=\textwidth]{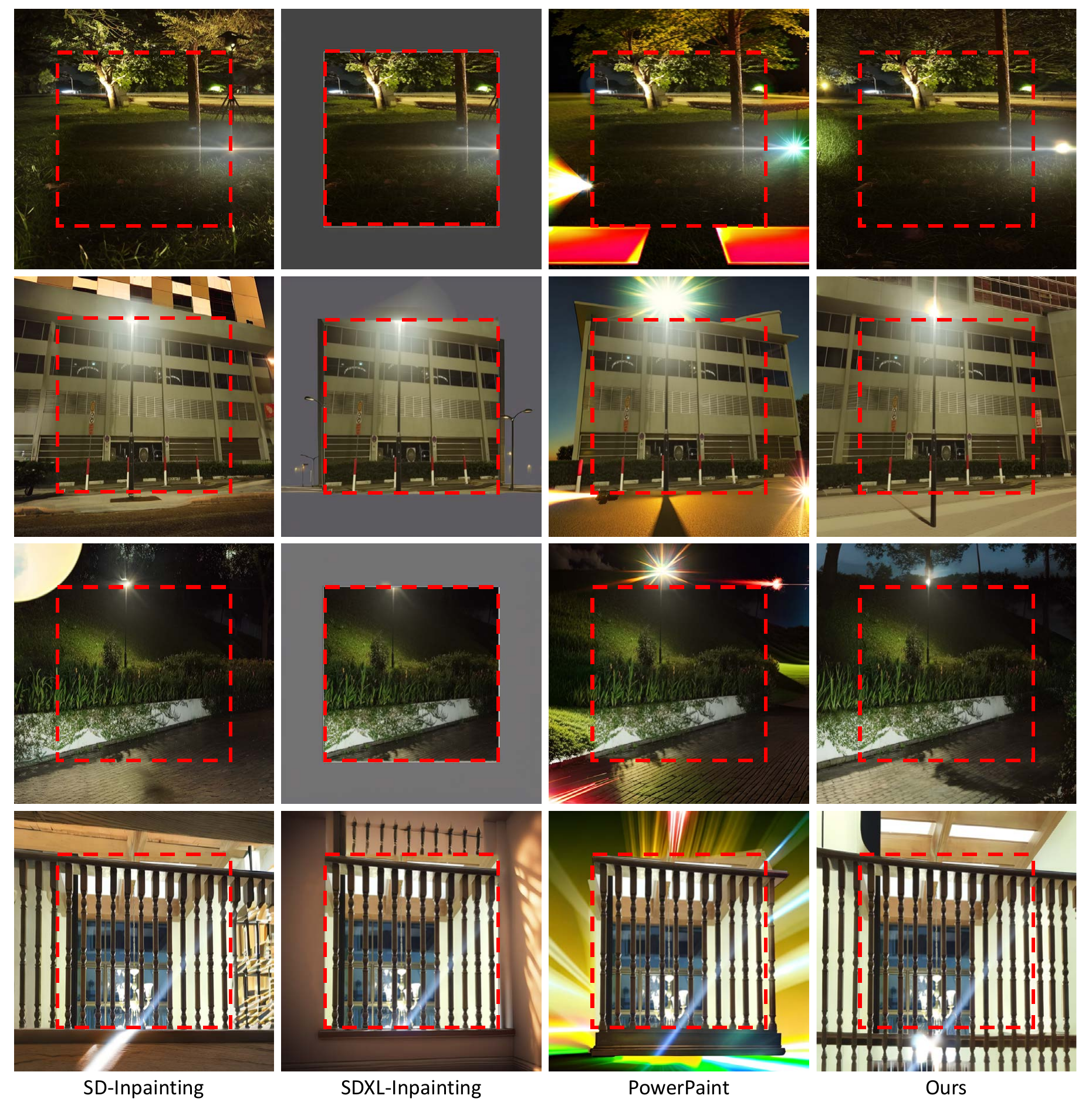}
    \caption{\textbf{Additional Qualitative Comparisons.} .}
    \label{fig:additional_qualitative_6}
\end{figure*} \fi

\end{document}